\documentclass{article}

\usepackage[margin=1in]{geometry}

\usepackage[T1]{fontenc}
\usepackage[numbers,square,sort]{natbib}
\usepackage{hyperref}
\hypersetup{
  colorlinks=false,
  pdfborder={0 0 1},
  citebordercolor={0 1 0},
  linkbordercolor={1 0 0},
  urlbordercolor={0 1 1}
}
\usepackage{url}
\usepackage{booktabs}
\usepackage{amsfonts}
\usepackage{amsmath}
\usepackage{microtype}
\usepackage{xcolor}
\usepackage{graphicx}
\usepackage{float}
\usepackage[section]{placeins}
\usepackage{caption}
\usepackage{multirow}
\usepackage{siunitx}
\usepackage{orcidlink}

\title{\bfseries CETUS: How Far Do Representations\\Trained on Earth Transfer to Cassini SAR of Titan?}

\author{%
  Kevin Lee{\hypersetup{pdfborder={0 0 0}}\orcidlink{0009-0004-0388-9260}} \\
  University of California at Los Angeles, Los Angeles, CA \\
  NASA Jet Propulsion Laboratory, California Institute of Technology, Pasadena, CA \\
  {\normalsize\texttt{kevinlee69720@g.ucla.edu}}
}

\date{}

\begin{document}

\maketitle

\begingroup
\renewcommand{\thefootnote}{}
\footnotetext[0]{Preprint.}
\endgroup

\begin{abstract}
\noindent
Cassini synthetic aperture radar (SAR) images reveal the dunes, plains, and lake basins of Titan,
providing an instance of representations learned from Earth imagery for planetary terrain
classification. Cross-domain Evaluation of Earth-to-Titan Transfer Using SAR (CETUS) compares
features from DINOv2, DOFA and CROMA with classical image measurements and features from an
untrained vision transformer on the U.S. Geological Survey's Cassini SAR mosaic. The classifiers
learn terrain labels from an expert geomorphological map and predict those labels in geographically
separate Titan regions. Under logistic regression settings, pretrained encoders achieve higher mean
macro F1 than the combined classical features. Encoder rankings change when feature scaling,
optimization, and regularization change together. Further training on Titan improves DINOv2
performance, degrades DOFA performance, and leads to mixed results for CROMA under the tested
settings. Architectural and input processing differences prevent these comparisons from isolating
the effect of pretraining. Classifier fitting and performance on individual terrain classes matter
when assessing representation transfer for planetary mapping. Since the map draws partly on the same
radar observations, the scores measure agreement with expert interpretation.
\end{abstract}

\section{Introduction}

Cassini synthetic aperture radar (SAR) images reveal Titan's plains, dune fields, uplands, labyrinth
terrain, lake basins and impact structures \citep{elachi2004cassini,lopes2020global}. Scientists
combine their shapes and radar brightness with complementary observations to reconstruct the moon's
surface history. Image classifiers could help search this extensive collection. CETUS\footnote{Tile
catalogs, predictions and evaluation code: \url{https://github.com/magnaprog/CETUS}.} tests this
possibility with classifiers that learn broad terrain labels from an expert map in one set of
regions and predict those map classes from SAR images elsewhere.

\begin{figure}[htbp]
\centering
\includegraphics[width=\linewidth]{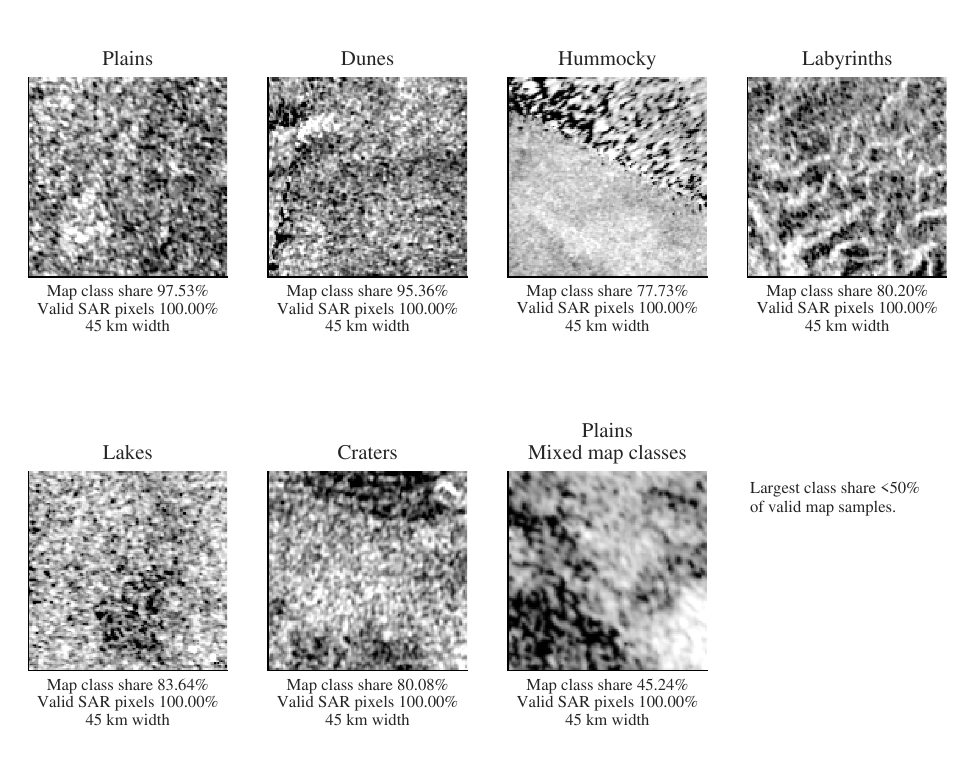}
\caption{Titan test images and their map labels. For each class, the example is the test tile whose
modal map fraction is closest to that class's median across the test set; tile ID breaks ties. The
mixed example follows the same rule among test tiles whose modal class occupies less than half the
valid map samples. Each image covers 45 km by 45 km. Map and SAR percentages use separate validity
masks. Brightness spans the 2nd to 98th percentile of each image's finite positive display DN.
Because each image has its own stretch, brightness differences between subplots have no common
radiometric scale. The examples illustrate the images and label mixtures produced by this sampling
rule; a geological assessment of their representativeness would require expert interpretation.}
\label{fig:terrain_examples}
\end{figure}

Planetary image classification already supports image search. For Mars,
\citet{wagstaff2021deployment} evaluated classifiers and deployed them for this purpose. Titan
presents a related task, with uneven radar coverage and terrain units that differ greatly in extent.
Accurate predictions for widespread plains can coexist with frequent errors on smaller units. A
class label for a large image area also provides less detail than a geologist's account of an
individual landform's shape, position and origin.

Pretraining on terrestrial imagery may provide useful visual features despite differences in Titan's
surface, observing geometry and image processing. How well a classifier uses those features affects
their evaluation. Poor fitting can obscure information in a representation, while an overall mean
score can conceal opposing changes among terrain classes. CETUS examines classifier fitting and
performance for individual classes using the same Titan mosaic and geomorphological map.

CETUS evaluates DINOv2, a general visual encoder pretrained on natural images; DOFA, which learns
across Earth observation sensors; and CROMA, whose pretraining pairs terrestrial radar and optical
imagery \citep{oquab2024dinov2,xiong2024dofa,fuller2023croma}. Each result describes a complete
model configuration, including its architecture, inputs and pretraining. Representation transfer
here means predicting map classes in other Titan regions using either the existing image features or
encoder weights trained further with Titan labels. The first comparison fits a classifier to each
pretrained representation, features from the Untrained ViT and four families of classical image
features using the same classifier settings. The next compares two ways of fitting a classifier with
the encoder weights unchanged, followed by training the final two transformer layers. A separate
experiment replaces some training tiles with tiles nearer the validation or test regions while
keeping the number of training examples in each class constant.

These experiments use an expanded Titan dataset that includes valid western image areas omitted by
the original map sampler. The sampler mishandled longitudes on a map spanning 0 to 360 degrees east.
Training and testing in the expanded dataset are separated geographically. The CETUS repository
provides the tile catalog, geographic divisions, predictions and reproduction instructions. Earlier
experiments trained classifiers on Earth and included an unlabeled Venus comparison.
Section~\ref{sec:earth_venus_comparison} examines them separately because their populations, labels
and radar products differ.

\section{Titan images and terrain labels}
\label{sec:titan_data}

\begin{figure}[htbp]
\centering
\includegraphics[width=\linewidth]{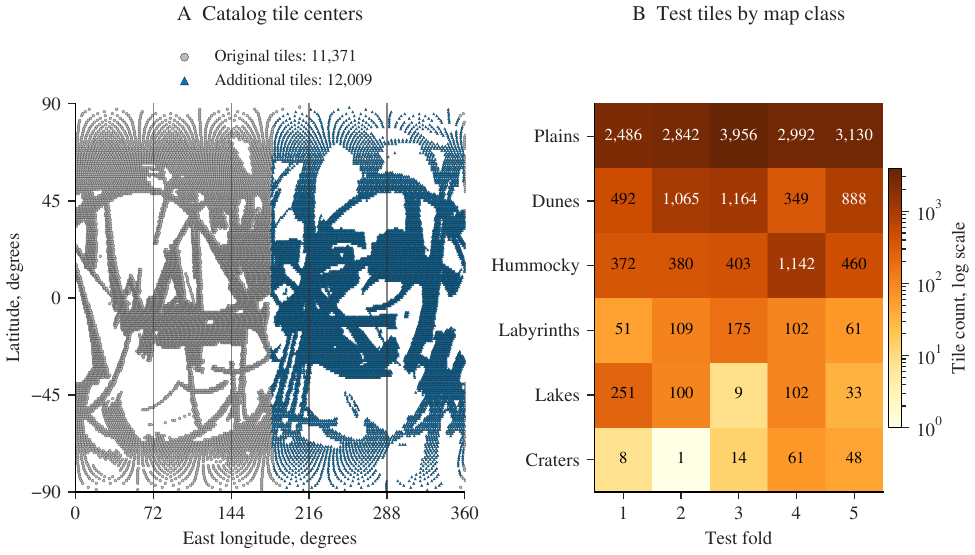}
\caption{Geographic coverage and numbers of test tiles. Left: original and additional tile centers
on a longitude and latitude grid. Vertical lines delimit the five sectors; markers show image
centers. Right: tile counts by map class and fold, with a logarithmic color scale. The Craters class
has one tile in fold 2 and 61 in fold 4, although the two folds receive equal weight in the primary
scores. Tile counts measure image sampling and may include several images of the same landform.}
\label{fig:titan_geography}
\end{figure}

The input tiles sample the U.S. Geological Survey (USGS) Cassini SAR and HiSAR display mosaic
\citep{usgs2015titan}. Each image tile contains $128\times128$ pixels over a nominal 45
km by 45 km area in a local metric projection. The mosaic's digital numbers, or DN, result from an 8
bit logarithmic stretch of calibrated source images. The tiles contain resampled values from that
mosaic. Tiles are retained when finite positive SAR values occupy at least 80\% of their pixels.
These DN describe image brightness; obtaining physical radar backscatter would require radiometric
information beyond this stretch. Pixel spacing determines sampling of the mosaic, while effective
resolution and observing geometry vary among its constituent images.

The map of \citet{lopes2020global} supplies six terrain labels: Plains, Dunes, Hummocky, Labyrinths,
Lakes and Craters. The benchmark uses Hummocky for the map's Mountains layer and Lakes for its
Basins layer. Hummocky includes mountains and degraded mountains; Lakes includes filled and empty
lakes and seas; Craters includes rims, ejecta, central peaks and fills. A
$64\times64$ grid samples the map over each tile, and the most frequent valid class
becomes its label. Ties go to the lowest class index. Label assignment retains every tile with a
mapped class, even when that class occupies less than half of the valid samples. Its fraction
describes the mixture of map classes in the image area, separately from confidence in the expert
interpretation. Map samples and SAR pixels use different validity masks and may cover different
portions of a tile.

Label assignment uses the first edition of the digital map polygons \citep{schoenfeld2024map}. At a
scale of 1:20{,}000{,}000, the global map simplifies regional mapping by omitting units smaller than
approximately 30~km$^2$ \citep{lopes2020global}. Other Cassini observations
inform the interpretation, especially outside SAR coverage. The sampling grid therefore represents
terrain divisions whose location and detail depend on that map. The Dunes class includes linear
dunes and sand sheets; the Lakes class can identify either liquid or an empty basin. A classifier
learns the predominant mapped unit in an image area, subject to these limits of scale and
interpretation.

Figure~\ref{fig:terrain_examples} shows the SAR images together with their modal map labels. The
class fractions make the choice reproducible. Whether each example is geologically representative
still requires expert assessment.

The expanded catalog contains 23{,}380 tiles. It preserves all original image arrays, adds 12{,}009
tiles and changes 15 labels on existing tiles. Independent inverse projection and direct lookup of
map pixels agree with the new sampler over the complete grid of possible image centers.
Figure~\ref{fig:titan_geography} shows where the additional images lie and how class counts vary
among regions. Since SAR also informed the expert map, agreement with its labels measures
reproduction of that interpretation.

\section{Experimental design}
\label{sec:experiment_design}

\subsection{Geographic separation and image normalization}
Titan is divided into five longitude sectors delimited by 0, 72, 144, 216, 288 and 360 degrees east.
Sectors are numbered from 1 to 5 in increasing longitude. Fold $f$ tests on
sector $f$, uses the next sector cyclically for validation and trains on the
remaining three. The Selk holdout contains 65 images within $5^\circ$ of
$6^\circ$N, $161^\circ$E. A conservative minimum distance of
108.640 km between centers of images in different sets gives a nominal 45 km separation between
their square image areas. This excludes another 69 tiles near Selk, validation tiles near the test
set and training tiles near the original validation or test sets. The primary split uses the
remaining tiles without replacements. Geographic separation reduces proximity between the sets;
spatial dependence can persist beyond the imposed distance.

The five test sets contain 23{,}246 distinct tiles. Training sets contain 12{,}348 to 14{,}644 tiles
and validation sets contain 3{,}453 to 5{,}420; all contain the six classes. In each fold, image
normalization uses up to 200 training tiles and 512 finite positive pixels per tile, with random
seed zero. DN are scaled between the 1st and 99th sample percentiles and clipped to
$[0,1]$. Only training pixels determine these limits. Every feature family uses
the same initial DN scaling for a given fold.

To test sensitivity to training geography, the replacement experiment reinstates all eligible
training tiles removed by the separation rule. For each class, it removes as many retained tiles as
it reinstates. Both classifiers use the same five training sets in each fold. Each set changes
2.20\% to 3.96\% of the training images in a fold. Each replacement set retains the 200 tiles that
determine image normalization, so the transform stays the same. Validation and test tiles, training
class counts and Selk exclusions also stay the same. For each of four encoders, logistic regression
recalculates feature standardization and fits a classifier on each replacement set. The stochastic
gradient descent (SGD) classifier uses five training seeds per set. This gives 100 additional
logistic regression fits and 500 SGD classifiers. Section~\ref{sec:training_geography} describes the
tile counts and the resulting score differences.

\subsection{Image representations and classifiers}
Each encoder is a vision transformer (ViT) with 768 output features. DINOv2 and the Untrained ViT
use the classification token, the representation assigned to the whole image. DOFA averages the
image patch representations, excluding that token, and applies a final normalization layer. DINOv2
ViT-B/14 repeats SAR into three channels and applies ImageNet channel normalization. DOFA ViT-B/16
uses one channel and numeric band identifier 13.78. The upstream Sentinel-1 example standardizes its
channels and uses 5.405~\citep{dofareadme2026}, whereas Appendix D of \citet{dofapreprintv2} assigns
SAR the identifier 3.75. CETUS omits the example's standardization. How these choices relate to the
model's pretraining inputs remains uncertain. CROMA ViT-B/8 duplicates SAR into two channels,
averages patch representations and passes the mean through its learned feedforward network. Both
channels contain the same polarization information. Bilinear resizing gives
$224\times224$ inputs, except for CROMA's $120\times120$.

The Untrained ViT baseline uses one ViT-B/14 with a single input channel and random seed 42. Its
encoder receives no training: the supervised classifier learns Titan terrain labels from the
features produced by its random weights. All classifier fits reuse this encoder, so the five SGD
repeats measure variation in classifier training for that one set of encoder weights. Its
single-channel input stem and lack of ImageNet channel normalization differ from DINOv2's input
processing. The comparison therefore combines pretraining with input differences and cannot isolate
the contribution of pretraining.

DOFA's final feature normalization retains its initial scale of one and bias of zero because the
checkpoint lacks those parameters. The layer still normalizes each feature vector. The checkpoint
also contains parameters for a separate normalization layer that the evaluated path does not use.
These missing and unused normalization parameters motivate checking the original input convention
before interpreting DOFA's performance or trying another training configuration.

For the primary comparison, each feature is standardized by its training mean and population
standard deviation (SD) before multinomial logistic regression (LR) is fitted. The classifier uses
an intercept, inverse class frequency weights, L2 regularization, $C=1$,
LBFGS, tolerance $10^{-4}$ and at most 2{,}000 iterations. The weights give each
class equal total weight in the loss. Each feature family has one fit per fold; validation data do
not choose among fits. Classical features comprise 13 intensity statistics, a 32-bin histogram,
1{,}820 texture features and their 1{,}865-dimensional concatenation. The texture measurements use
cooccurrence matrices, local binary patterns, Gabor filters, histograms of oriented gradients and
wavelets. Histogram and texture calculations also rescale each tile between its own 1st and 99th
percentiles. Classifier settings are the same across families; feature definitions and dimensions
differ.

The combined classical features provide the primary baseline for assessing whether an encoder adds
predictive value beyond conventional image measurements. A separate majority predictor assigns every
image to the most numerous class in the training fold. It uses class frequency alone, while
classifiers using classical features also use image information.

The second linear classifier uses unstandardized encoder features and SGD. Its cross-entropy loss
has inverse class frequency weights. The classifier is trained for 100 epochs with learning rate
0.01, momentum 0.9, zero weight decay, cosine decay and batch size 256, then evaluated at the final
epoch for five classifier seeds in each fold. The common logistic regression settings were chosen
after inspecting these SGD results, but before examining the logistic regression or combined
classical results. The comparison changes feature scaling, optimization and regularization together.

A nearest-neighbor classifier provides a further comparison. It takes an unweighted vote among the
20 closest training features by cosine similarity and assigns ties to the lowest class index.

\subsection{Training part of the encoder on Titan}
For each pretrained model, training updates the final two transformer layers and a new linear
classifier for all 50 epochs, with five optimization seeds per fold. The 75 fits use AdamW, learning
rates of $10^{-4}$ for the encoder and $10^{-3}$ for the
classifier, weight decay 0.01, cosine decay, microbatches of eight and accumulation over four
microbatches. The loss uses inverse class frequency weights, and training uses automatic mixed
precision. Image normalization, model inputs and geographic divisions stay the same. Evaluation uses
the earliest epoch attaining the highest validation macro recall among epochs with optimizer
updates. The new classifier has an independent initialization from the earlier linear classifiers.
Appendix~\ref{app:encoder_training} gives the training counts and checks of the saved models.

Table~\ref{tab:classifier_settings} compares the three ways of fitting models. All use inverse class
frequency weights and the same image normalization for a given fold.

\begin{table}[htbp]
\centering
\caption{Settings for classifier fitting and partial encoder training. Standardization centers and
scales features using the training mean and population SD.}
\label{tab:classifier_settings}
\begingroup
\fontsize{9}{11}\selectfont
\setlength{\tabcolsep}{4pt}
\renewcommand{\arraystretch}{1.12}
\begin{tabular}{@{}p{0.17\linewidth}p{0.24\linewidth}p{0.22\linewidth}p{0.29\linewidth}@{}}
\toprule
\raggedright Fitting choice & \raggedright Common logistic regression & \raggedright SGD classifier & \raggedright Partial fine-tuning \tabularnewline
\midrule
\raggedright Model parameters fitted & \raggedright Linear classifier & \raggedright Linear classifier & \raggedright Final two transformer layers and a new linear classifier \tabularnewline
\addlinespace
\raggedright Standardize classifier inputs & \raggedright Training mean and population SD & \raggedright None & \raggedright None \tabularnewline
\addlinespace
\raggedright Optimizer and regularization & \raggedright LBFGS with L2, $C=1$ & \raggedright SGD, zero weight decay & \raggedright AdamW, weight decay 0.01 \tabularnewline
\addlinespace
\raggedright Training duration & \raggedright At most 2,000 iterations & \raggedright 100 epochs & \raggedright All 50 epochs \tabularnewline
\addlinespace
\raggedright Model evaluated & \raggedright One fit; validation data unused & \raggedright Classifier from the final epoch & \raggedright Earliest epoch attaining the highest validation macro recall among epochs with optimizer updates \tabularnewline
\addlinespace
\raggedright Fits per fold & \raggedright One per feature family & \raggedright Five classifier seeds per encoder & \raggedright Five optimization seeds per pretrained encoder \tabularnewline
\bottomrule
\end{tabular}
\endgroup

\end{table}

\subsection{Performance measures and regional averages}
For each of the six classes, precision, recall and F1 are calculated, with undefined ratios set to
zero. Macro scores are the arithmetic means of these class scores; in particular, F1 is calculated
before averaging classes. For SGD and encoder training, five seeds are averaged in each fold,
followed by the mean and sample SD of the five fold values. Differences involving encoder training
compare the same folds, using the same seed for the SGD comparison and one logistic regression fit
per fold. For encoder training, validation macro recall determines the epoch; the test results
include all three metrics. The SD measures variation among these regions. Overlapping training sets
can induce dependence among cross-validation scores \citep{bengio2004variance}. Spatial dependence
among Titan images is a further concern when treating these folds as independent geographic
repetitions.

The primary average gives each region equal influence for every class, even when class counts
differ. Scores are also calculated after combining confusion counts from all test folds separately
for each encoder-training seed, then averaging the resulting seed scores. Logistic regression
contributes one matrix across all test tiles. This second calculation weights the available test
population differently. Both calculations concern the sampled mosaic, with uneven geographic
coverage and potentially several images per landform. Every primary comparison uses regional
averages. The alternative weighting, distance checks and map polygon analysis were designed after
inspecting the main results and use the existing predictions. The experiment replacing training
tiles was specified before its results were examined.

Bold numbers identify the highest displayed mean within the comparison named in each table caption.
Ties at the displayed precision receive equal emphasis. Bold type does not establish statistical
significance.

\enlargethispage{\baselineskip}
\section{Results and error interpretation}

\subsection{Pretrained features and classical image measurements}

\begin{table}[!htbp]
\centering
\caption{Logistic regression applied to eight feature families, with encoder weights unchanged.
Training features determine standardization; inverse class frequency weights give every class equal
total weight. Precision, recall and F1 are macro averages in percent, followed by their sample SD
across five folds. All 40 fits finish in fewer than 2{,}000 iterations without warnings. Bold
indicates the highest mean in each metric column across all eight feature families.}
\label{tab:feature_comparison}
\small
\begin{tabular}{lrrr}
\toprule
Feature family & Precision & Recall & F1 \\
\midrule
DINOv2 & $35.49 \pm 2.94$ & $43.05 \pm 3.85$ & $36.81 \pm 2.76$ \\
DOFA & $35.24 \pm 2.97$ & $43.12 \pm 5.40$ & $35.91 \pm 3.52$ \\
CROMA & $\mathbf{38.46} \pm 2.62$ & $\mathbf{47.23} \pm 5.77$ & $\mathbf{39.34} \pm 2.48$ \\
Untrained ViT & $30.80 \pm 4.98$ & $34.87 \pm 3.93$ & $27.28 \pm 3.96$ \\
Intensity statistics & $30.46 \pm 2.59$ & $39.29 \pm 6.83$ & $23.62 \pm 1.76$ \\
Intensity histogram & $25.60 \pm 2.12$ & $32.89 \pm 6.99$ & $18.67 \pm 1.58$ \\
Texture & $27.64 \pm 1.11$ & $32.84 \pm 3.00$ & $28.02 \pm 1.40$ \\
Combined classical & $33.51 \pm 3.84$ & $39.53 \pm 5.06$ & $33.67 \pm 2.75$ \\
\bottomrule
\end{tabular}
\end{table}

\begin{figure}[!htbp]
\centering
\includegraphics[width=\linewidth]{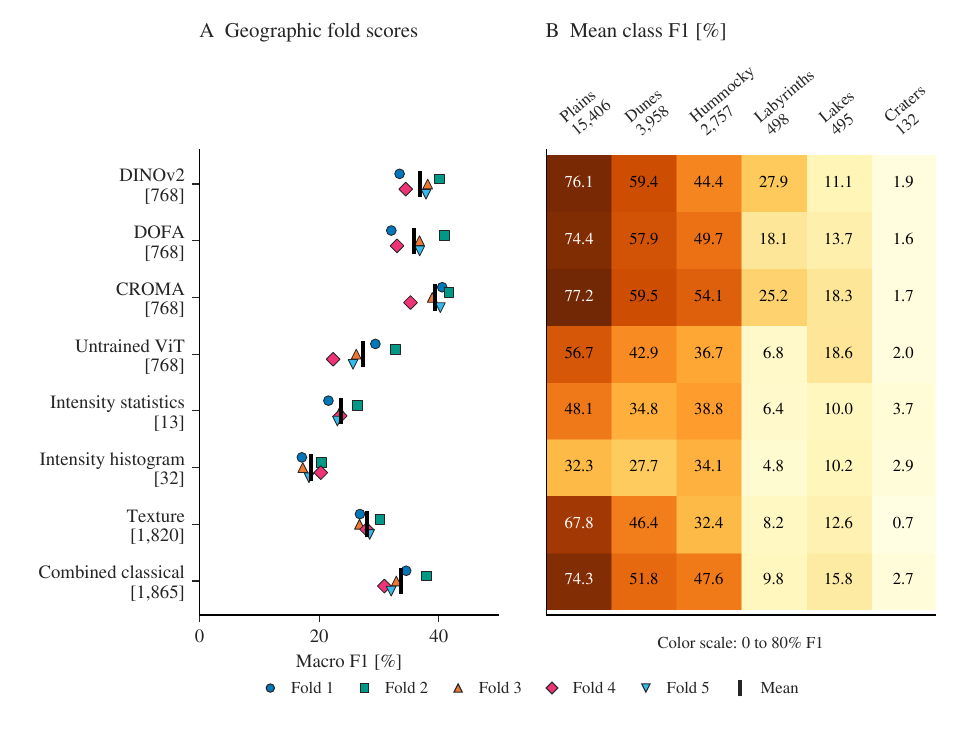}
\caption{Logistic regression with each feature family. Left: five geographic fold scores and their
mean. Right: mean F1 for each class, with the number of distinct test tiles above each column. All
families use the same training and test images and classifier settings. They share the initial DN
scaling, but subsequent input transformations differ. Numbers beside family names give feature
dimensions. A common color scale spans 0 to 80\% F1. Training sets overlap between folds. The
Craters class contributes between one and 61 tiles per fold, which limits comparisons among regional
class scores.}
\label{fig:feature_comparison}
\end{figure}

CROMA has the highest mean macro F1 in Table~\ref{tab:feature_comparison} and
Figure~\ref{fig:feature_comparison}. It exceeds the combined classical baseline in every fold and by
5.67 percentage points on average. DINOv2 and DOFA exceed that baseline by 3.14 and 2.24 points in
mean F1. Combining intensity and texture improves mean F1 over either alone. Classifier settings and
images are the same; input transformations, dimensions and architectures differ. Further tuning
could change the ranking.

\FloatBarrier

\subsection{Classifier fitting changes the encoder ranking}
\begin{figure}[ht]
\centering
\includegraphics[width=\linewidth]{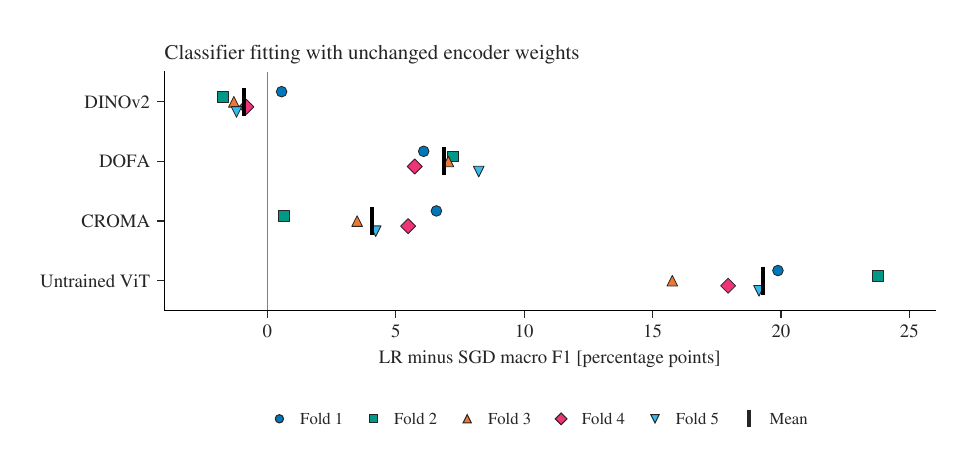}
\caption{Change in macro F1 when logistic regression replaces SGD, with identical encoder features
and image sets. Colored markers show the difference for each geographic fold in percentage points;
black marks give the mean. The SGD score first averages five training seeds. Logistic regression
changes feature scaling, optimization, regularization and the training schedule together, so the
comparison assesses the fitting procedures as a whole.}
\label{fig:classifier_comparison}
\end{figure}

With SGD, DINOv2 leads at 37.71\% F1, followed by CROMA at 35.26\% and DOFA at 29.04\%. Logistic
regression puts CROMA first and raises the Untrained ViT result from 7.97\% to 27.28\%, an increase
of 19.3 percentage points. For the Untrained ViT, the SGD classifiers assign at least 95.77\% of
each test set to one class; logistic regression predicts all six classes in every fold. Both fits
learn from Titan labels and use the same encoder features. The higher LR score shows that these
features permit better predictions than the nearly constant SGD outputs. Scaling, optimization and
regularization change together, leaving their individual contributions unresolved. All pretrained
encoders still exceed this particular random initialization in every fold with logistic regression.
This comparison with one encoder initialization does not establish the advantage over random
encoders in general. Figure~\ref{fig:classifier_comparison} shows the change for each region.

Both classifiers include an intercept. In exact arithmetic, an invertible affine rescaling of
features leaves the set of possible linear classifications intact. Rescaling changes numerical
conditioning and the meaning of an L2 penalty in the original feature coordinates. Fitting choices
can therefore change performance even when the representation stays the same. Identifying the
contribution of scaling would require varying it separately from optimization and regularization.

\FloatBarrier

\begin{table}[ht]
\centering
\caption{SGD classifiers and nearest neighbors on the same geographic test sets and encoder
features. Values give mean macro precision, recall and F1 in percent, with sample SD across five
folds. SGD metrics first average the training seeds in each fold. Bold indicates the highest mean in
each metric column, separately for SGD and nearest neighbors.}
\label{tab:sgd_and_neighbors}
\small
\begin{tabular}{lrrr}
\toprule
Encoder & Precision & Recall & F1 \\
\midrule
\multicolumn{4}{l}{SGD classifier} \\
DINOv2 & $\mathbf{36.02} \pm 3.51$ & $44.70 \pm 4.30$ & $\mathbf{37.71} \pm 3.56$ \\
DOFA & $32.57 \pm 3.76$ & $45.25 \pm 7.67$ & $29.04 \pm 2.94$ \\
CROMA & $35.85 \pm 3.73$ & $\mathbf{48.45} \pm 7.40$ & $35.26 \pm 4.05$ \\
Untrained ViT & $6.01 \pm 2.17$ & $16.66 \pm 0.004$ & $7.97 \pm 2.47$ \\
\midrule
\multicolumn{4}{l}{Nearest neighbors} \\
DINOv2 & $\mathbf{48.89} \pm 8.91$ & $27.62 \pm 3.79$ & $28.86 \pm 5.04$ \\
DOFA & $42.00 \pm 7.57$ & $26.79 \pm 2.48$ & $28.57 \pm 2.78$ \\
CROMA & $48.63 \pm 11.85$ & $\mathbf{28.89} \pm 2.52$ & $\mathbf{30.83} \pm 2.90$ \\
Untrained ViT & $21.45 \pm 7.85$ & $17.06 \pm 0.51$ & $14.99 \pm 0.76$ \\
\bottomrule
\end{tabular}
\end{table}

The performance measure also changes which encoder leads: DINOv2 has the highest SGD F1, while CROMA
has the highest SGD recall in Table~\ref{tab:sgd_and_neighbors}. For all three pretrained encoders,
SGD has higher mean macro recall and lower mean macro precision than nearest neighbors. Its mean F1
exceeds nearest neighbors by 8.85 points for DINOv2, 0.47 for DOFA and 4.43 for CROMA. Thus
classifier choice changes both the encoder ranking and the tradeoff between precision and recall on
these images.

\subsection{How training on Titan changes performance}
\begin{figure}[htbp]
\centering
\includegraphics[width=\linewidth]{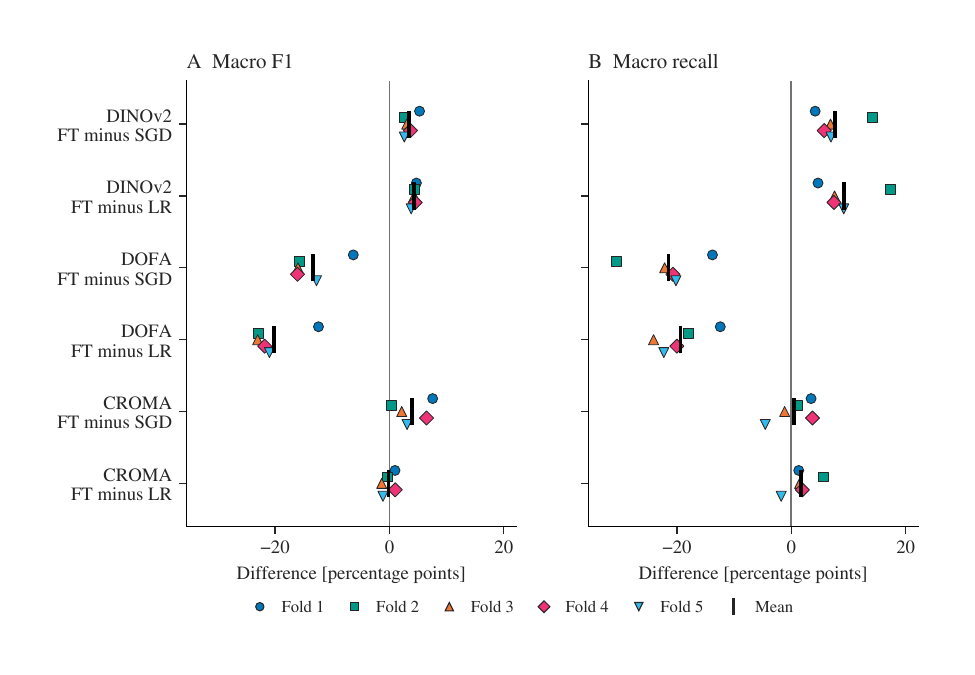}
\caption{Training the last two transformer layers, compared with training only a classifier on the
same geographic divisions. FT denotes partial fine-tuning. Left: differences in test macro F1.
Right: differences in test macro recall. Differences subtract the indicated classifier scores from
the FT scores and are expressed in percentage points. Colored symbols show five fold values; black
marks give their mean. FT and SGD first average five training seeds per fold, with corresponding
seeds compared for the difference. Logistic regression has one fit per fold. Validation macro recall
determines the FT epoch. Optimizer and training settings also differ, so these comparisons assess
the full training configurations.}
\label{fig:encoder_training}
\end{figure}

Training the last two layers raises DINOv2 to $41.16\pm2.60$\% macro F1. This exceeds
logistic regression by $4.35\pm0.34$ percentage points and SGD by
$3.45\pm1.10$ points, with both improvements present in every fold. DOFA reaches
$15.68\pm3.35$\%. Its differences from the two classifiers are
$-20.23\pm4.45$ and $-13.36\pm4.17$ points, respectively, and both are
negative in every fold. CROMA reaches $39.17\pm2.31$\%, improving on SGD by
$3.91\pm3.03$ points in the mean and in every fold. Against logistic regression, its
difference is $-0.17\pm1.12$ points, with two positive and three negative fold
differences. Figure~\ref{fig:encoder_training} shows both comparisons. CROMA's macro recall
nevertheless increases by $1.70\pm2.62$ points relative to logistic regression.
Whether encoder training improves performance therefore depends on both the comparison classifier
and the performance measure.

\FloatBarrier

\begin{table}[ht]
\centering
\caption{Partial fine-tuning (FT) compared with classifier training alone. LR denotes logistic
regression. Scores are percentages; differences are FT less the comparison score, in percentage
points. Cells give the mean and sample SD of five folds after averaging five FT seeds. SGD
differences compare corresponding seeds; LR has one fit per fold. Validation macro recall determines
the FT epoch. Bold indicates the highest mean fine-tuning score in each metric column. Differences
from earlier fits remain in ordinary type.}
\label{tab:encoder_training_scores}
\small
\setlength{\tabcolsep}{4pt}
\begin{tabular}{lrrrr}
\toprule
Encoder & FT F1, \% & FT recall, \% & FT minus LR F1, pp & FT minus SGD F1, pp \\
\midrule
DINOv2 & $\mathbf{41.16} \pm 2.60$ & $\mathbf{52.28} \pm 7.05$ & $4.35 \pm 0.34$ & $3.45 \pm 1.10$ \\
DOFA & $15.68 \pm 3.35$ & $23.76 \pm 3.61$ & $-20.23 \pm 4.45$ & $-13.36 \pm 4.17$ \\
CROMA & $39.17 \pm 2.31$ & $48.93 \pm 4.52$ & $-0.17 \pm 1.12$ & $3.91 \pm 3.03$ \\
\bottomrule
\end{tabular}

\end{table}

\subsection{Which terrain classes improve?}
\begin{figure}[htbp]
\centering
\includegraphics[width=\linewidth]{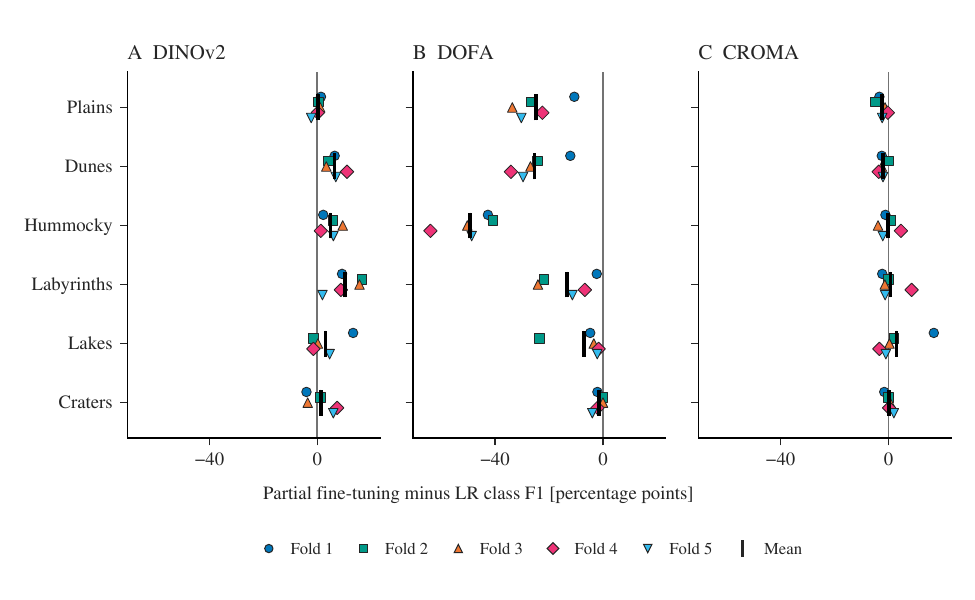}
\caption{Change in class F1 after training the last two transformer layers, relative to logistic
regression on their original features. Each colored symbol gives one fold's difference after
averaging five training seeds. Black marks give the mean across folds. Positive values indicate
closer agreement with the modal map class. The common horizontal scale places DOFA's broad decline
beside the smaller changes for DINOv2 and CROMA. Several images can represent the same physical
landform.}
\label{fig:class_f1_changes}
\end{figure}

Relative to logistic regression, DINOv2 improves Dunes, Hummocky and Labyrinths F1 in every fold,
with little change in mean Plains F1. CROMA gains on some classes and loses on others. DOFA's
decline spans all six class means. Figure~\ref{fig:class_f1_changes} shows the regional variation
behind these averages. Explaining why the encoders respond differently would require experiments
that vary the training choices separately.

DINOv2's largest mean F1 gains are 10.37 percentage points for Labyrinths, 6.33 for Dunes and 4.90
for Hummocky. CROMA gains 2.94 points for Lakes while losing 2.44 for Plains and 2.13 for Dunes.
DOFA's Hummocky F1 falls from 49.71\% to 0.42\%, and its recall falls from 61.82\% to 0.24\%.
Hummocky contains 2{,}757 test tiles. DOFA's deterioration therefore extends to a class represented
by many test images.

CROMA's small change in macro F1 also depends on how regions are weighted. The difference is
$-0.17$ percentage points when each fold receives equal weight and
$+0.85$ when confusion counts from all test folds are combined separately for
each training seed. With the latter calculation, DINOv2 still improves and DOFA still declines.
Table~\ref{tab:regional_weighting} compares the two calculations; the primary results continue to
give every region equal weight.

\begin{table}[ht]
\centering
\caption{Effect of regional weighting on the comparison of FT with LR. FT denotes partial
fine-tuning and LR logistic regression. F1 scores are percentages; differences are percentage
points. The primary calculation averages the five regional scores equally. The second calculation
combines confusion counts across test folds for each seed before calculating F1, then averages
seeds.}
\label{tab:regional_weighting}
\small
\begin{tabular}{lrrrr}
\hline
Model & Primary $FT-LR$ & All tiles LR & All tiles FT & All tiles $FT-LR$ \\
\hline
DINOv2 & $4.35$ & $38.43$ & $42.99$ & $4.55$ \\
DOFA & $-20.23$ & $37.68$ & $16.21$ & $-21.47$ \\
CROMA & $-0.17$ & $40.98$ & $41.82$ & $0.85$ \\
\hline
\end{tabular}

\end{table}

\begin{table}[ht]
\centering
\caption{Class precision (P), recall (R) and F1 for partial fine-tuning (FT) and logistic regression
(LR), in percent. Class ratios are calculated for each seed and fold; FT seeds are averaged in each
fold, then the five folds receive equal weight. LR has one fit per fold. SD measures variation among
folds; the tile count includes each test image once. Undefined ratios are zero. Several tiles may
depict portions of the same landform. Bold indicates the higher mean F1 between FT and LR for each
encoder and class.}
\label{tab:encoder_training_classes}
\small
\setlength{\tabcolsep}{3pt}
\begin{tabular}{@{}llrrrrrrr@{}}
\toprule
Encoder & Class & Tiles & FT P, \% & FT R, \% & FT F1, \% & LR P, \% & LR R, \% & LR F1, \% \\
\midrule
DINOv2 & Plains & 15406 & \shortstack{$88.05$\\$\pm 2.69$} & \shortstack{$68.03$\\$\pm 7.48$} & \shortstack{$\mathbf{76.24}$\\$\pm 4.32$} & \shortstack{$85.22$\\$\pm 1.36$} & \shortstack{$68.92$\\$\pm 5.70$} & \shortstack{$76.10$\\$\pm 3.65$} \\
DINOv2 & Dunes & 3958 & \shortstack{$61.80$\\$\pm 7.60$} & \shortstack{$71.52$\\$\pm 9.70$} & \shortstack{$\mathbf{65.76}$\\$\pm 7.47$} & \shortstack{$53.23$\\$\pm 11.12$} & \shortstack{$68.04$\\$\pm 8.21$} & \shortstack{$59.43$\\$\pm 9.37$} \\
DINOv2 & Hummocky & 2757 & \shortstack{$46.21$\\$\pm 3.87$} & \shortstack{$57.96$\\$\pm 18.87$} & \shortstack{$\mathbf{49.34}$\\$\pm 7.55$} & \shortstack{$37.36$\\$\pm 5.34$} & \shortstack{$57.65$\\$\pm 17.63$} & \shortstack{$44.43$\\$\pm 8.22$} \\
DINOv2 & Labyrinths & 498 & \shortstack{$30.10$\\$\pm 10.64$} & \shortstack{$63.81$\\$\pm 12.99$} & \shortstack{$\mathbf{38.23}$\\$\pm 7.48$} & \shortstack{$25.24$\\$\pm 9.92$} & \shortstack{$38.88$\\$\pm 18.40$} & \shortstack{$27.85$\\$\pm 5.32$} \\
DINOv2 & Lakes & 495 & \shortstack{$12.78$\\$\pm 9.88$} & \shortstack{$32.84$\\$\pm 23.53$} & \shortstack{$\mathbf{14.10}$\\$\pm 11.58$} & \shortstack{$10.51$\\$\pm 8.82$} & \shortstack{$20.87$\\$\pm 20.69$} & \shortstack{$11.09$\\$\pm 12.33$} \\
DINOv2 & Craters & 132 & \shortstack{$4.64$\\$\pm 6.43$} & \shortstack{$19.54$\\$\pm 23.64$} & \shortstack{$\mathbf{3.27}$\\$\pm 3.13$} & \shortstack{$1.37$\\$\pm 1.95$} & \shortstack{$3.93$\\$\pm 5.70$} & \shortstack{$1.94$\\$\pm 2.67$} \\
DOFA & Plains & 15406 & \shortstack{$68.04$\\$\pm 4.56$} & \shortstack{$42.45$\\$\pm 9.90$} & \shortstack{$49.67$\\$\pm 8.27$} & \shortstack{$85.97$\\$\pm 0.55$} & \shortstack{$65.75$\\$\pm 5.09$} & \shortstack{$\mathbf{74.42}$\\$\pm 3.33$} \\
DOFA & Dunes & 3958 & \shortstack{$24.07$\\$\pm 10.25$} & \shortstack{$61.79$\\$\pm 4.48$} & \shortstack{$32.49$\\$\pm 10.96$} & \shortstack{$52.99$\\$\pm 10.96$} & \shortstack{$65.97$\\$\pm 12.28$} & \shortstack{$\mathbf{57.90}$\\$\pm 8.50$} \\
DOFA & Hummocky & 2757 & \shortstack{$1.87$\\$\pm 4.19$} & \shortstack{$0.24$\\$\pm 0.53$} & \shortstack{$0.42$\\$\pm 0.94$} & \shortstack{$44.34$\\$\pm 6.07$} & \shortstack{$61.82$\\$\pm 22.25$} & \shortstack{$\mathbf{49.71}$\\$\pm 8.77$} \\
DOFA & Labyrinths & 498 & \shortstack{$9.38$\\$\pm 3.30$} & \shortstack{$10.78$\\$\pm 12.26$} & \shortstack{$4.84$\\$\pm 2.91$} & \shortstack{$13.83$\\$\pm 8.26$} & \shortstack{$33.10$\\$\pm 14.88$} & \shortstack{$\mathbf{18.13}$\\$\pm 8.21$} \\
DOFA & Lakes & 495 & \shortstack{$11.38$\\$\pm 17.47$} & \shortstack{$27.29$\\$\pm 23.42$} & \shortstack{$6.66$\\$\pm 5.68$} & \shortstack{$12.97$\\$\pm 12.01$} & \shortstack{$27.98$\\$\pm 23.72$} & \shortstack{$\mathbf{13.73}$\\$\pm 14.33$} \\
DOFA & Craters & 132 & \shortstack{$0.00$\\$\pm 0.00$} & \shortstack{$0.00$\\$\pm 0.00$} & \shortstack{$0.00$\\$\pm 0.00$} & \shortstack{$1.33$\\$\pm 1.39$} & \shortstack{$4.08$\\$\pm 5.36$} & \shortstack{$\mathbf{1.61}$\\$\pm 1.66$} \\
CROMA & Plains & 15406 & \shortstack{$87.71$\\$\pm 1.54$} & \shortstack{$65.63$\\$\pm 4.72$} & \shortstack{$74.78$\\$\pm 2.85$} & \shortstack{$87.21$\\$\pm 1.60$} & \shortstack{$69.49$\\$\pm 5.79$} & \shortstack{$\mathbf{77.22}$\\$\pm 3.55$} \\
CROMA & Dunes & 3958 & \shortstack{$55.28$\\$\pm 12.92$} & \shortstack{$61.92$\\$\pm 9.98$} & \shortstack{$57.36$\\$\pm 9.16$} & \shortstack{$56.14$\\$\pm 8.32$} & \shortstack{$64.67$\\$\pm 12.85$} & \shortstack{$\mathbf{59.49}$\\$\pm 8.07$} \\
CROMA & Hummocky & 2757 & \shortstack{$48.69$\\$\pm 5.71$} & \shortstack{$65.13$\\$\pm 20.27$} & \shortstack{$53.82$\\$\pm 7.53$} & \shortstack{$48.63$\\$\pm 5.55$} & \shortstack{$65.68$\\$\pm 19.00$} & \shortstack{$\mathbf{54.14}$\\$\pm 5.66$} \\
CROMA & Labyrinths & 498 & \shortstack{$18.80$\\$\pm 5.90$} & \shortstack{$54.09$\\$\pm 12.98$} & \shortstack{$\mathbf{25.86}$\\$\pm 5.14$} & \shortstack{$19.21$\\$\pm 6.34$} & \shortstack{$44.94$\\$\pm 17.30$} & \shortstack{$25.16$\\$\pm 4.78$} \\
CROMA & Lakes & 495 & \shortstack{$19.64$\\$\pm 18.36$} & \shortstack{$40.64$\\$\pm 26.14$} & \shortstack{$\mathbf{21.27}$\\$\pm 19.59$} & \shortstack{$18.41$\\$\pm 16.92$} & \shortstack{$33.87$\\$\pm 20.57$} & \shortstack{$18.33$\\$\pm 13.98$} \\
CROMA & Craters & 132 & \shortstack{$2.43$\\$\pm 3.66$} & \shortstack{$6.20$\\$\pm 8.37$} & \shortstack{$\mathbf{1.95}$\\$\pm 2.15$} & \shortstack{$1.14$\\$\pm 1.13$} & \shortstack{$4.73$\\$\pm 5.18$} & \shortstack{$1.71$\\$\pm 1.60$} \\
\bottomrule
\end{tabular}

\end{table}

\subsubsection{Craters remains difficult to predict}
Every feature family has mean Craters F1 below 4\% with logistic regression. For pretrained
encoders, it ranges from 1.61\% to 1.94\%, with precision from 1.14\% to 1.37\%. Better macro F1 can
accompany worse prediction of this small class. For example, changing DOFA's classifier from SGD to
logistic regression raises macro F1 from 29.04\% to 35.91\%, while Craters recall falls from 40.85\%
to 4.08\% and Craters precision from 1.99\% to 1.33\%. Assessing predictions of crater terrain
therefore requires the class scores as well as the overall score.

After encoder training, mean Craters F1 is 3.27\% for DINOv2, 0.00\% for DOFA and 1.95\% for CROMA.
DINOv2 reaches 4.64\% Craters precision and 19.54\% recall. Thus, despite its higher macro F1, most
of its Craters predictions disagree with the map. Table~\ref{tab:encoder_training_classes} gives
precision, recall and F1 for every class and encoder.

DINOv2's Craters F1 becomes 5.15\% when confusion counts are combined across test folds for each
seed. The finding of F1 below 4\% refers specifically to equal regional weights. Combining counts
across seeds or weighting fold F1 by test size would answer still different statistical questions.
These low scores, together with the scale of the map labels, leave practical crater detection to be
assessed separately.

The CROMA confusion matrix in Table~\ref{tab:croma_confusion} shows how predictions differ from map
labels. Of 132 tiles labeled Craters by the map, 99 receive Hummocky and seven receive Craters.
CROMA predicts Craters for 484 tiles in total. Of 495 Lakes tiles, 157 receive Dunes and 125 receive
Lakes. Each test image contributes once to these counts, so they weight regions differently from the
primary class ratios. The matrix identifies frequent disagreements between map labels and
predictions; interpreting the structures or materials in those images requires geological
inspection.

\begin{table}[htbp]
\centering
\caption{CROMA predictions from logistic regression across all five test folds. Each of 23,246 test
tiles appears once. Rows give modal map classes and columns give predictions. These counts weight
the test population by tile count, whereas the primary class scores weight the five regions equally.
CROMA is the encoder with the highest mean F1 in the common classifier comparison.}
\label{tab:croma_confusion}
\begingroup
\fontsize{9}{11}\selectfont
\setlength{\tabcolsep}{4pt}
\renewcommand{\arraystretch}{1.12}
\begin{tabular}{@{}lrrrrrrr@{}}
\toprule
& \multicolumn{6}{c}{Predicted map class} & \\
\cmidrule(lr){2-7}
Map class & Plains & Dunes & Hummocky & Labyrinths & Lakes & Craters & Test tiles \\
\midrule
Plains & 10,689 & 1,687 & 1,771 & 656 & 364 & 239 & 15,406 \\
Dunes & 875 & 2,660 & 141 & 110 & 117 & 55 & 3,958 \\
Hummocky & 425 & 87 & 1,965 & 76 & 42 & 162 & 2,757 \\
Labyrinths & 143 & 57 & 70 & 203 & 14 & 11 & 498 \\
Lakes & 98 & 157 & 76 & 29 & 125 & 10 & 495 \\
Craters & 21 & 0 & 99 & 1 & 4 & 7 & 132 \\
\midrule
Prediction total & 12,251 & 4,648 & 4,122 & 1,075 & 666 & 484 & 23,246 \\
\bottomrule
\end{tabular}
\endgroup

\end{table}

The five folds contain 8, 1, 14, 61 and 48 Craters test tiles. The fold containing one tile
contributes one fifth of mean Craters recall. With one fit per fold, changing that tile from
incorrect to correct raises mean six-class macro recall across folds by 3.33 percentage points. The
relation between these tile counts and distinct impact structures requires a separate inventory: the
Craters class includes rims, ejecta and fills, while an image assigned another modal class may still
include some crater terrain. Both scale and class mixture matter when interpreting an individual
disagreement.

\subsubsection{Map concentration and validation sensitivity}
Two map polygon components account for 82 of 132 Craters test tiles, as shown in
Table~\ref{tab:crater_map_components}. A component can include several features, and a physical
feature may extend beyond an image whose modal label is Craters. Lakes test counts also vary
substantially among regions, with 251, 100, 9, 102 and 33 test tiles.

\begin{table}[ht]
\centering
\caption{Craters test tiles and the map polygon components they represent. The largest component
column identifies the map component with the most test tiles; the last column gives that count. A
separate geological inventory is needed to relate these components to distinct impact structures.}
\label{tab:crater_map_components}
\small
\begin{tabular}{rrrrr}
\hline
Fold & Craters tiles & Components & Largest component & Tiles in largest \\
\hline
1 & 8 & 2 & 18 & 7 \\
2 & 1 & 1 & 17 & 1 \\
3 & 14 & 5 & 14 & 8 \\
4 & 61 & 4 & 58 & 52 \\
5 & 48 & 5 & 53 & 30 \\
\hline
\end{tabular}

\end{table}

Fold 1 contains one Craters validation tile. Switching its prediction between correct and incorrect
changes six-class validation macro recall by $100/6\approx16.67$ percentage points. At the
epochs used for evaluation, three DINOv2 seeds, one DOFA seed and no CROMA seed classify that tile
correctly. Thus both correct and incorrect predictions occur among the models chosen by validation
recall. The saved training histories contain only overall metrics. Evaluating a different rule for
choosing the epoch would require class predictions for every epoch.

\subsection{Sensitivity to training geography}
\label{sec:training_geography}
\begin{figure}[ht]
\centering
\includegraphics[width=\linewidth]{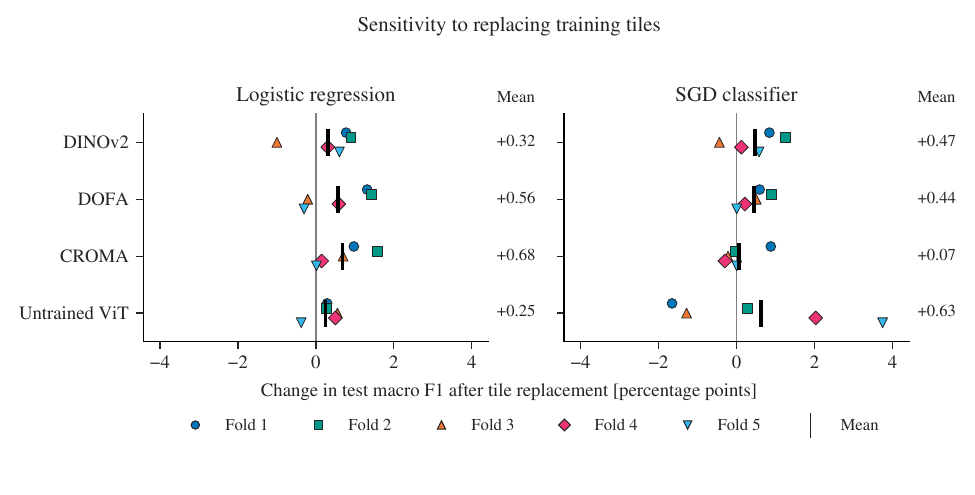}
\caption{Effect of replacing training tiles while keeping class counts and test sets constant. Left:
logistic regression. Right: SGD. Colored points give the change in macro F1 for each fold, in
percentage points; black marks and printed values give the mean. LR averages five replacement sets,
each compared with the original fit. Each set reinstates all eligible excluded tiles; the sets vary
which retained tiles are removed. SGD compares corresponding training seeds, averages five training
seeds per set, then averages five sets. Initial DN scaling stays the same; LR recalculates feature
standardization. All fits contribute, and every LR fit finishes without a warning or reaching its
iteration limit. Locations and image content both change, and training sets overlap between folds.}
\label{fig:training_geography}
\end{figure}

Replacing some training tiles changes macro F1 by $0.32\pm0.77$,
$0.56\pm0.81$, $0.68\pm0.63$ and $0.25\pm0.37$ percentage
points for DINOv2, DOFA, CROMA and the Untrained ViT with logistic regression. The corresponding
changes with SGD are $0.47\pm0.65$, $0.44\pm0.35$,
$0.07\pm0.47$ and $0.63\pm2.27$ points. These values average five
replacement sets per fold; SGD first averages its training seeds for each set. The SD describes the
five fold differences. All 100 logistic fits finish without warnings or reaching the iteration
limit. Figure~\ref{fig:training_geography} shows each fold. The replacements affect 2.20\% to 3.96\%
of training tiles while retaining class counts. Because they change both location and image content,
the results measure sensitivity to those replacements. Isolating the effect of proximity would
require a different experiment.

The folds reinstate 486, 486, 429, 528 and 322 tiles; their original training sets contain 14{,}603,
12{,}542, 12{,}348, 13{,}350 and 14{,}644 tiles. Four folds reinstate no Craters tiles and one
reinstates two, limiting what this experiment can reveal about crater terrain. Fold differences
include both positive and negative values, except for CROMA with LR and DOFA with SGD, which improve
in all five folds. Table~\ref{tab:training_counts} gives the class counts. The experiment changes
training geography only through these small replacements. Larger changes in spatial separation could
behave differently.

\begin{table}[ht]
\centering
\caption{Training, validation and test counts for the encoder training and tile replacement
experiments. Replaced tiles are those reinstated near another region; in each class, the same number
of tiles is removed as reinstated. Percentages divide the replacement count by the original training
size. The last three columns count Craters tiles. Validation and test sets stay the same. Every
count refers to tiles, independent of the number of seeds or replacement sets.}
\label{tab:training_counts}
\small
\setlength{\tabcolsep}{4pt}
\begin{tabular}{rrrrrrr}
\toprule
Fold & \shortstack{Training\\tiles} & \shortstack{Replaced\\tiles} & \shortstack{Training\\\%} & \shortstack{Reinstated\\Craters} & \shortstack{Validation\\Craters} & \shortstack{Test\\Craters} \\
\midrule
1 & 14{,}603 & 486 & 3.33 & 0 & 1 & 8 \\
2 & 12{,}542 & 486 & 3.87 & 2 & 14 & 1 \\
3 & 12{,}348 & 429 & 3.47 & 0 & 59 & 14 \\
4 & 13{,}350 & 528 & 3.96 & 0 & 48 & 61 \\
5 & 14{,}644 & 322 & 2.20 & 0 & 8 & 48 \\
\bottomrule
\end{tabular}
\end{table}

The overall F1 comparison covers all test images, including those far from the reinstated training
images. Only 1{,}136 of 23{,}246 test centers lie within the original minimum separation of at least
one reinstated training center, representing 3.33\% to 6.72\% of each fold. Error changes in these
nearby images take both signs. Tables~\ref{tab:nearby_test_tiles}
and~\ref{tab:errors_by_training_distance} compare them with the remaining test images. Retraining
can also change predictions farther away, so this geographic comparison complements the overall
score.

\subsubsection{Test images near reinstated training images}
A test image is classified as nearby when the distance along Titan's spherical surface from its
center to at least one reinstated training center is strictly smaller than the original minimum
separation. The calculation uses the same Titan radius and coordinates as the geographic split. Some
reinstated tiles are near validation images and far from test images, because the original
separation rule considered both sets. Table~\ref{tab:nearby_test_tiles} gives the nearby test
counts. Center distances describe proximity; image overlap and statistical dependence would require
additional checks.

\begin{table}[ht]
\centering
\caption{Test images near reinstated training images. The training column counts reinstated tiles.
The nearby test column counts distinct test centers within the original minimum separation of at
least one reinstated training center. Percentages use the full test count of each fold.}
\label{tab:nearby_test_tiles}
\small
\begin{tabular}{rrrrr}
\hline
Fold & Reinstated & Test & Nearby & Nearby (\%) \\
\hline
1 & 486 & 3660 & 207 & 5.66 \\
2 & 486 & 4497 & 155 & 3.45 \\
3 & 429 & 5721 & 301 & 5.26 \\
4 & 528 & 4748 & 319 & 6.72 \\
5 & 322 & 4620 & 154 & 3.33 \\
\hline
\end{tabular}

\end{table}

\begin{table}[ht]
\centering
\caption{Change in test error rate after replacing training tiles for logistic regression, in
percentage points. Each cell gives the change among nearby test tiles followed by the change among
remaining tiles. Positive values mean more disagreement with the map. Values average five
replacement sets, and the table includes every encoder and fold.}
\label{tab:errors_by_training_distance}
\small
\begin{tabular}{rcccc}
\hline
Fold & DINOv2 (E/R) & DOFA (E/R) & CROMA (E/R) & Untrained ViT (E/R) \\
\hline
1 & $-2.51 / +0.00$ & $-2.42 / -0.91$ & $-0.48 / -0.19$ & $+0.97 / -1.22$ \\
2 & $-1.94 / -0.12$ & $-3.10 / -0.49$ & $-2.45 / -0.71$ & $+0.26 / +1.07$ \\
3 & $-1.33 / +0.24$ & $+1.06 / -0.01$ & $-4.32 / -0.44$ & $-2.19 / -1.04$ \\
4 & $+1.69 / -0.56$ & $-1.50 / -0.42$ & $-2.51 / -0.05$ & $-0.94 / -0.56$ \\
5 & $-4.94 / +0.04$ & $-4.94 / +0.66$ & $-4.03 / +0.43$ & $+1.04 / +0.44$ \\
\hline
\end{tabular}

\end{table}

Only one fold contains nearby Craters test images, and it contains two. The accompanying results
give class counts and errors for each replacement set. With so few Craters examples and simultaneous
changes in training locations and image content, these differences leave the appropriate separation
distance and the magnitude of any score inflation unresolved.

\subsection{Map mixture and geographic variation with SGD}
SGD predictions from the geographic evaluation allow errors to be compared by map composition, image
coverage and region. Repeating these analyses for logistic regression or encoder training would
require separate calculations.

\paragraph{Mixed terrain:}
The mixture analysis groups test images by map class and fold, then compares images whose modal map
fraction is below 0.75 with those whose fraction is at least 0.90. Both groups must contain at least
20 tiles. Twenty of 30 class and fold combinations qualify, all outside Craters. After averaging
classifier errors for each tile and weighting these combinations equally, the more mixed images have
19.04, 11.01 and 15.92 percentage points more errors for DINOv2, DOFA and CROMA. The thresholds were
set before this analysis, after inspecting the overall scores. Mixed terrain, image conditions and
local context can all contribute to the difference. This association motivates reporting class
fractions alongside prediction scores. Confidence in the map remains a separate question.

\paragraph{Region and training repeats:}
For the pretrained encoders, macro F1 varies more among geographic folds than among training seeds
in a given fold. Fold SDs are 3.56, 2.94 and 4.05 percentage points for DINOv2, DOFA and CROMA; the
mean seed SDs within folds are 0.225, 0.110 and 0.061 points. Training images and class counts also
differ among folds, so these quantities describe two kinds of variation without separating their
causes. The Untrained ViT behaves differently: each SGD classifier assigns 95.77\% to 100\% of test
tiles to a single class. Its F1 of 7.97\% falls below the 13.27\% mean obtained by always predicting
Plains, the most numerous training class. This collapse occurs for this random encoder with the
stated SGD settings; the logistic regression results show that a different fit can use its features
more effectively.

\begin{figure}[ht]
\centering
\includegraphics[width=\linewidth]{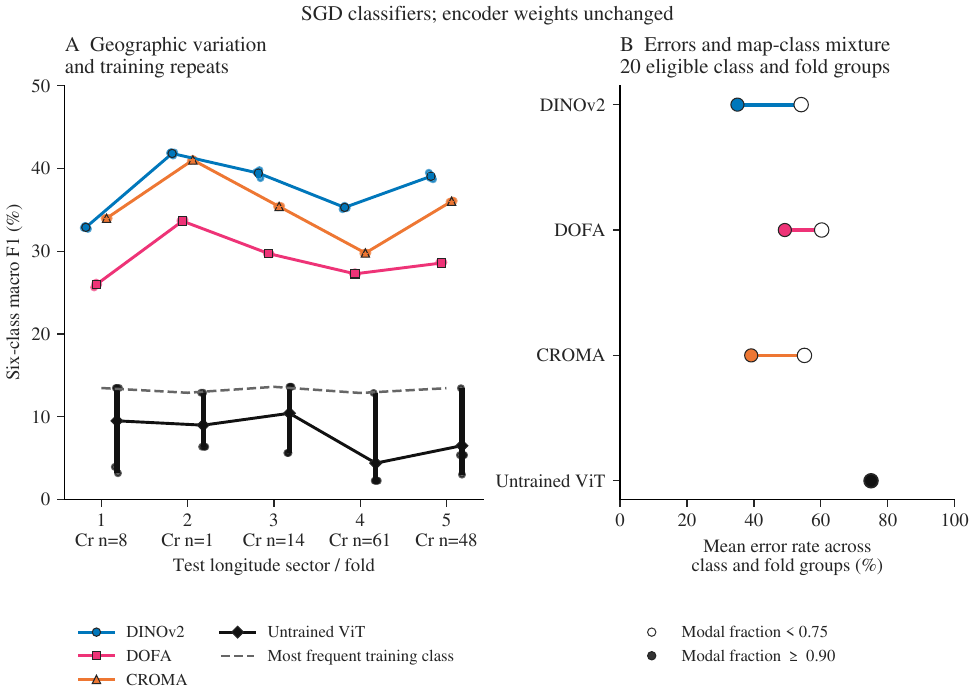}
\caption{SGD performance by region, training repeat and map mixture. Left: F1 for five training
seeds in each fold, with connected means and the full seed ranges. The dashed baseline always
predicts the most numerous training class. Right: error rates for modal map fractions below 0.75 and
at least 0.90, across 20 eligible class and fold combinations. Each group contains at least 20 tiles
per combination, and combinations receive equal weight. Craters has too few examples to enter this
comparison. Untrained ViT's markers almost coincide because its predictions are nearly constant. The
plotted differences may include effects of both terrain mixture and image conditions.}
\label{fig:sgd_regions_and_map_mixture}
\end{figure}

\paragraph{Where the mixture comparison applies:}
The eligible groups contain 5{,}804 images with modal fractions below 0.75 and 13{,}206 with
fractions at or above 0.90. Plains, Dunes and Hummocky enter in all five folds, Labyrinths in three
and Lakes in two. More mixed images have higher error in 17 of 20 groups for DINOv2, 15 for DOFA and
17 for CROMA. Nearby images may share terrain or observing conditions, so these sign counts describe
the sampled groups without providing independent repetitions of a causal test.
Figure~\ref{fig:sgd_confusion} shows the class assignments, and
Figure~\ref{fig:sgd_regions_and_map_mixture} displays variation with geography, training repeats and
mixture.

A separate comparison groups images by valid SAR fraction, below 0.95 or at least 0.99, with the
same minimum number of tiles. Thirteen class and fold combinations qualify. Mean error differences
are +3.87, +2.17 and $-1.09$ percentage points for DINOv2, DOFA and CROMA. Lower
valid SAR coverage accompanies higher error for two encoders and slightly lower error for the third.
Both analyses retain the catalog's minimum SAR validity of 0.80 and examine mixture and coverage
separately. Direct image inspection and an analysis of their combined effects would be needed to
distinguish the contributions of each.

\paragraph{Errors in neighboring images:}
The spatial analysis calculates correlations in SGD errors between tiles of the same map class and
fold, using center distances greater than zero and at most 45 km. Raw correlations are positive in
21 of 24 eligible groups for DINOv2, 23 of 25 for DOFA and 24 of 25 for CROMA. This indicates
spatial structure in test errors. The smallest Craters groups contain too few images, and constant
error values make correlation undefined in some other groups. Dependence remains unassessed in the
omitted groups. Appendix~\ref{app:errors_by_distance} gives the distance intervals, weighting and
minimum counts. Measuring whether proximity to training images inflates a score, and by how much,
would require an additional experiment.

\FloatBarrier

\section{Classifiers trained on Earth and a comparison with Venus}
\label{sec:earth_venus_comparison}
The earlier experiments train classifiers on Earth site labels and test them against Titan map
classes. They use the original Titan catalog and spatial groups without separation zones. The
current experiments instead train classifiers with Titan labels and test their geographic
generalization. Table~\ref{tab:experiment_comparison} distinguishes the two datasets and designs.
The comparisons below equalize training counts and class composition or vary image preprocessing.
Differences in label quality, geographic sampling and radar products still complicate the
interpretation of transfer between Earth and Titan.

\begin{table}[ht]
\centering
\caption{The two datasets and experimental designs. Each score refers to the images and training
settings in its column. Several aspects change between the experiments, so a difference in score
alone cannot identify the effect of any one change.}
\label{tab:experiment_comparison}
\small
\begin{tabular}{p{0.17\linewidth}p{0.37\linewidth}p{0.37\linewidth}}
\toprule
Property & Current geographic evaluation & Earlier Earth transfer experiments \\
\midrule
Titan catalog & Expanded catalog, 23{,}380 tiles & Original catalog, 11{,}371 tiles \\
Geography & Five longitude sectors with separation between training, validation and testing & Spatial groups without separation zones \\
Prediction task & Six modal Titan map classes & Six classes within Titan; five for Earth transfer \\
Fitting & Logistic regression, SGD and training of the final two encoder layers & Original classifier settings and additional comparisons \\
Other domains & Earth and Venus considered separately & Earth source labels require curation; Venus is unlabeled \\
Interpretation & Geographic map agreement and sensitivity to fitting & Results for the original sample and additional comparisons \\
\bottomrule
\end{tabular}
\end{table}

The original Earth split contains 352 training tiles and 46 validation tiles. Sentinel-1 VV GRD
products are converted to physical backscatter, linear $\sigma^0$, with ESA SNAP,
then to dB before normalization. Titan images contain display DN. Sampling both at 45 km by 45 km
with the same output spacing leaves differences in native resolution, point-spread functions,
incidence angles and speckle. Label curation and geographic sampling also differ.
Appendix~\ref{app:original_dataset} describes the images and their geographic divisions.

\subsection{Equal training counts and alternative image preprocessing}
\label{sec:earth_preprocessing}

The experiments compare training counts and image preprocessing on the original catalog using five
output classes: Plains, Dunes, Labyrinths, Lakes and Craters. Both experiments exclude the 16
Egyptian images incorrectly named Wadi Rum and evaluate the same 10{,}292 Titan test tiles,
including 17 Craters. They retain the original longitude omission and geographic groups without
separation zones. The remaining Earth labels still need curation, including 42 Arctic image areas
with no mapped land. DOFA uses band identifier 5.405 for Earth and 13.78 for Titan; the original
classifier features used 13.78 for both. This follows the repository's example convention, whose
relationship to pretraining remains uncertain. Several other settings also change, leaving the
identifier's individual effect unknown.

The first experiment gives each training domain 334 images, with class counts of 109, 174, 29, 16
and 6. Five subsets are formed using groups of Earth sites and radar products and Titan spatial
groups. Equal class counts still permit differences in geographic diversity. Earth normalization
uses the catalog moments for each subset, and Titan normalization uses that subset's training DN
percentiles. The Earth and Titan classifiers in each comparison use the same Titan image transform.
Titan labels determine the Titan subset, so the experiment uses information from Titan's training
set. Each classifier is trained for 100 SGD epochs and 200 updates, with inverse frequency weights,
learning rate 0.01, momentum 0.9, zero weight decay and cosine decay. The last epoch is evaluated
for five training seeds. Training on Titan exceeds training on Earth in F1 by 24.98, 16.50 and 24.51
percentage points for DINOv2, DOFA and CROMA. These differences persist after equalizing counts,
while their physical and sampling causes remain to be separated.

\begin{table}[t]
\centering
\small
\caption{Equal training counts by domain and class in the original dataset. Macro precision, recall
and F1 are percentages with sample SD across five fold means. Each fold first averages five training
seeds for each subset, then gives the five subsets equal weight. Domains use identical class counts
and numbers of training updates. Undefined class scores are zero. Training sets overlap among folds
and subsets, limiting inferences that require independent repetitions. Bold indicates the higher
mean between Earth and Titan training for each encoder and metric.}
\label{tab:earth_training_sets}
\begin{tabular}{llrrr}
\toprule
Encoder & Classifier training & Precision & Recall & F1 \\
\midrule
DINOv2 & Earth & $24.76\pm1.70$ & $25.14\pm0.46$ & $18.31\pm0.87$ \\
       & Titan & $\mathbf{41.54}\pm1.53$ & $\mathbf{53.25}\pm3.95$ & $\mathbf{43.29}\pm1.84$ \\
DOFA & Earth & $5.89\pm0.49$ & $22.99\pm1.44$ & $9.33\pm0.71$ \\
     & Titan & $\mathbf{32.50}\pm1.66$ & $\mathbf{57.65}\pm7.39$ & $\mathbf{25.83}\pm2.91$ \\
CROMA & Earth & $24.04\pm2.09$ & $31.03\pm7.45$ & $11.48\pm2.04$ \\
      & Titan & $\mathbf{37.03}\pm2.72$ & $\mathbf{66.23}\pm6.36$ & $\mathbf{35.99}\pm3.35$ \\
Untrained ViT & Earth & $\mathbf{3.15}\pm0.01$ & $\mathbf{20.00}\pm0.00$ & $\mathbf{3.85}\pm0.02$ \\
            & Titan & $\mathbf{3.15}\pm0.01$ & $\mathbf{20.00}\pm0.00$ & $\mathbf{3.85}\pm0.02$ \\
\bottomrule
\end{tabular}
\end{table}

The preprocessing experiment uses all 336 remaining Earth training tiles, with class counts of 109,
174, 29, 16 and 8. One sample of training pixels supplies both moment and percentile scaling limits.
Quantization and smoothing vary independently, while Titan features stay the same in a given fold.
Titan training pixels determine its normalization, and images from both training domains determine
the Earth smoothing scale. Every fit chooses $\sigma=0.65$ output pixels from the
alternatives specified in advance. Only training images enter these choices.

Five training seeds per condition use 100 SGD epochs, batch size 256 and the same optimizer settings
as the experiment with equal training counts; update traces are unavailable for these runs.
Corresponding seeds are compared across conditions. The baseline uses continuous moment scaling
without smoothing.

\begin{table}[t]
\centering
\small
\caption{Eight image preprocessing conditions for the original dataset. Macro F1 is in percent, with
sample SD of five fold means after averaging five training seeds. Moment scaling uses limits three
population SDs below and above the training dB mean; percentile scaling uses the 1st and 99th
percentiles. The 256 setting quantizes values before optional smoothing. Continuous retains values
without additional quantization. These transformations alter image statistics; physical calibration
and averaging of independent radar looks require information unavailable in the Titan DN. Bold
indicates the highest mean F1 in each encoder column across the displayed preprocessing conditions.}
\label{tab:earth_preprocessing}
\begin{tabular}{lllrrr}
\toprule
Scaling limits & Quantization & Smoothing & DINOv2 & DOFA & CROMA \\
\midrule
Moments & Continuous & Off & $17.65\pm0.92$ & $9.40\pm0.84$ & $11.42\pm2.02$ \\
Moments & Continuous & On & $12.71\pm0.68$ & $\mathbf{9.56}\pm1.94$ & $8.50\pm0.43$ \\
Moments & 256 & Off & $17.62\pm0.89$ & $9.40\pm0.84$ & $11.48\pm2.02$ \\
Moments & 256 & On & $12.73\pm0.69$ & $9.55\pm1.93$ & $8.51\pm0.43$ \\
Percentiles & Continuous & Off & $\mathbf{18.94}\pm0.93$ & $8.67\pm1.71$ & $11.62\pm2.74$ \\
Percentiles & Continuous & On & $13.26\pm0.75$ & $8.90\pm1.90$ & $\mathbf{12.84}\pm1.27$ \\
Percentiles & 256 & Off & $18.91\pm0.93$ & $8.65\pm1.71$ & $11.59\pm2.73$ \\
Percentiles & 256 & On & $13.23\pm0.76$ & $8.90\pm1.90$ & $12.83\pm1.28$ \\
\bottomrule
\end{tabular}
\end{table}

In Table~\ref{tab:earth_preprocessing}, smoothing lowers mean F1 for DINOv2 and raises it slightly
for DOFA under each scaling and quantization setting. For CROMA, smoothing lowers mean F1 with
moment scaling and raises it with percentile scaling. Both experiments use Titan training images to
estimate image statistics. Their conclusions apply to the original sample and the stated SGD fits.
Extending them to the expanded catalog or separating geographic and physical causes requires further
work. The CETUS reproduction guide identifies the complete class metrics, differences between
corresponding runs and numerical checks for these comparisons.

\subsection{Transfer using the full original training sets}
The Titan training folds contain 6{,}778 to 6{,}794 tiles. Both classifiers retain six output
logits, but transfer evaluation covers the five classes present in Earth training data, excluding
Hummocky. On those same five classes, classifiers trained on Titan attain macro recalls of 54.9\%,
57.6\% and 64.7\% for DINOv2, DOFA and CROMA. Table~\ref{tab:transfer_metrics} gives precision,
recall and F1 for classifiers trained on Earth. This comparison includes differences in sample
count, class composition, sites, sensors and image processing.

Precision and F1 are calculated from the original confusion matrices and verified against the saved
predictions. Table~\ref{tab:transfer_metrics} averages the five class scores for each seed, then the
seeds in each fold, then the folds. A zero denominator gives a score of zero; macro F1 averages
class F1 values. CROMA has the lowest macro F1 among the pretrained encoders despite its highest
recall. Its Craters precision is 0.57\% and F1 is 1.13\%, indicating frequent Craters assignments to
images with a different modal map label. Raw intensity statistics attain 23.76
$\pm$ 0.98\% macro F1 under the same averaging scheme.

\begin{table}[t]
\centering
\caption{Classifiers trained on Earth and evaluated on Titan for the five classes present in the
original Earth training set. Precision, recall and F1 are percentages from the original confusion
matrices. Values after $\pm$ give the sample SD of five fold means. DOFA uses
identifier 13.78 for both domains in these results. Tables~\ref{tab:earth_training_sets}
and~\ref{tab:earth_preprocessing} provide additional experiments with different training samples and
separately estimated image normalization. Bold indicates the highest mean in each metric column.}
\label{tab:transfer_metrics}
\begin{tabular}{lccc}
\toprule
\textbf{Model} & \textbf{Macro precision} & \textbf{Macro recall} & \textbf{Macro F1} \\
\midrule
DINOv2 & 23.84 & 25.08 & 18.61 $\pm$ 0.81 \\
DOFA & \textbf{26.11} & 25.45 & \textbf{18.76} $\pm$ 0.83 \\
CROMA & 24.77 & \textbf{32.32} & 10.17 $\pm$ 2.02 \\
\bottomrule
\end{tabular}
\end{table}

\subsection{Venus as an unlabeled distribution comparison}
\label{sec:venus_comparison}

The Venus comparison uses encoder features from 2{,}945 tiles of the Magellan SAR FMAP
mosaic~\citep{saunders1992magellan}, acquired at \SI{2.385}{\giga\hertz} in S-band. These images
lack terrain labels, so the comparison concerns feature distributions. The comparisons in
Table~\ref{tab:venus_comparison} draw equal numbers of images from each domain, using 398 Earth,
2{,}945 Venus and 11{,}306 Titan tiles outside Selk as the available populations. The input
description names the FMAP product from right-looking observations \citep{usgs2015venusmagellan}.
The raster header confirms one 8 bit band with 75 m spacing; look direction and global coverage
remain unverified from that header. Product documentation describes a quantized radar cross section
normalized by an angular scattering law. Feature extraction uses its DN encoding directly, without
conversion back to physical backscatter.

The comparison uses squared maximum mean discrepancy,
MMD$^2$~\citep{gretton2012kernel}. MMD measures the distance between kernel
mean embeddings. With a characteristic kernel such as the Gaussian radial basis function (RBF),
population MMD equals zero exactly when the distributions agree. Estimates from finite samples carry
sampling uncertainty. Even agreement between feature distributions would leave the physical
interpretation of the images unresolved.

\begin{samepage}
The comparison uses the squared biased estimator with an RBF kernel on feature samples of equal size
$X$ and $Y$:
\begin{equation}
\widehat{\operatorname{MMD}}_{\mathrm{b}}^2 =
\frac{1}{n^2}\sum_{i,j=1}^{n} k(x_i,x_j) +
\frac{1}{n^2}\sum_{i,j=1}^{n} k(y_i,y_j) -
\frac{2}{n^2}\sum_{i,j=1}^{n} k(x_i,y_j).
\end{equation}
\end{samepage}

Each table entry averages estimates from 100 repeated samples. Comparisons involving Earth use 398
images per domain; the Venus and Titan comparison uses 2{,}000 per domain. For a given encoder, all
three comparisons use the same bandwidth. Equal sample sizes within each pair do not remove possible
differences in finite-sample bias between comparisons. The estimates describe differences between
feature distributions. Relating them to classification performance, radar frequency, surface
composition, atmosphere, incidence angle and image processing requires separate experiments.

\begin{table}[H]
\centering
\caption{MMD$^2$ between planetary image domains, averaged over 100 repeated
samples with one RBF bandwidth per encoder. Each repeat uses 398 tiles per domain when Earth enters
the comparison and 2{,}000 for Venus and Titan. Each column uses a different feature space, so
numerical comparisons across encoders have no common scale. Earth and Titan give the smallest
MMD$^2$ for DINOv2, DOFA and CROMA. The images differ in radar products,
frequency, surface materials, atmosphere, acquisition and preprocessing, which prevents assigning
this ordering to one physical factor.}
\label{tab:venus_comparison}
\begin{tabular}{lccc}
\toprule
\textbf{Domains} & \textbf{DINOv2} & \textbf{DOFA} & \textbf{CROMA} \\
\midrule
Earth (C-band) and Venus (S-band) & 0.216 & 0.261 & 0.744 \\
Venus (S-band) and Titan (Ku-band) & 0.297 & 0.497 & 0.509 \\
Earth (C-band) and Titan (Ku-band) & 0.112 & 0.165 & 0.412 \\
\bottomrule
\end{tabular}
\end{table}

Earth and Titan give the smallest MMD$^2$ for all three pretrained encoders:
0.112 for DINOv2, 0.165 for DOFA and 0.412 for CROMA. Earth and Venus come next for DINOv2 and DOFA;
Venus and Titan come next for CROMA.

These estimates use a different sampling scheme from Table~\ref{tab:feature_distributions}. For the
Earth and Titan comparison here, all 398 Earth tiles and all Titan images outside Selk are
available, and 398 images are drawn from each of these two domains. Each encoder uses one bandwidth
for its three domain comparisons. The fold analysis instead samples 352 Earth training tiles and 352
Titan training tiles per repeat, estimates a new bandwidth each time, and averages five fold means.
DOFA is consequently 0.165 here and 0.183 in that calculation. The two values describe different
comparisons.

The ordering of domain pairs also differs from the ordering of radar frequency separations. Because
acquisition, processing, coverage and surface statistics change together, a physical explanation
requires additional observations. Testing Venus as an intermediate training domain would further
require an experiment in classification or encoder training.

\section{Discussion}

\subsection{Relation to previous work}

\paragraph{Geospatial foundation models:}
Pretraining on Earth observation data has produced several general image encoders.
DOFA~\citep{xiong2024dofa} generates patch embedding weights from band identifiers across sensing
modalities. CROMA~\citep{fuller2023croma} learns from Sentinel-1 radar and Sentinel-2 optical images
of the same locations, with losses for cross-modal agreement and image reconstruction.
DINOv2~\citep{oquab2024dinov2} learns from natural images through self-distillation and supplies a
general vision comparison. Section~\ref{sec:experiment_design} gives the inputs and encoder outputs
used here. SatMAE~\citep{cong2022satmae} and Prithvi~\citep{jakubik2024prithvi} extend masked
autoencoders to temporal and multispectral satellite imagery. These models establish the terrestrial
context for the present experiments, alongside a growing literature on planetary foundation models.

\paragraph{Planetary image analysis:}
Planetary machine learning spans terrain classification, object detection and segmentation.
\citet{burl1998volcanoes} trained classifiers to detect volcanoes in Magellan SAR images of Venus
three decades ago. Studies of Cassini SAR include fuzzy cognitive maps for Titan surface
analysis~\citep{furfaro2010cryovolcanism} and lake segmentation~\citep{bratsolis2010lakes}. Cloud
instance segmentation~\citep{yahn2025clouds} examines Titan's atmosphere through another imaging
modality. For Mars, \citet{wagstaff2018deep} trained AlexNet with ImageNet pretraining for two
tasks: classifying rover images and classifying cropped landmarks in orbital High Resolution Imaging
Science Experiment (HiRISE) images. AI4Mars~\citep{swan2021ai4mars} initially collected
approximately 326K human annotation submissions for ${\sim}35$K rover images,
combining multiple annotations per image for terrain segmentation aimed at autonomous driving.
Mars-Bench~\citep{purohit2025marsbench} evaluates foundation models on Mars science tasks, and
MOMO~\citep{purohit2026momo} learns from several Mars orbital sensors.

On Earth, PANGAEA~\citep{marsocci2024pangaea} evaluates geospatial foundation models for
segmentation, change detection and regression. CrossEarth~\citep{guo2025crossearth} and
CrossEarth-SAR~\citep{ye2026crossearthsar} study segmentation in terrestrial domains absent from
training, with the latter concentrating on radar. CETUS examines geographic prediction of Titan map
classes and considers the Earth and Venus experiments separately in
Section~\ref{sec:earth_venus_comparison}.

\paragraph{Radar interpretation and terrestrial analogs:}
Studies of individual Titan landforms combine morphology with physical constraints.
\citet{paillou2014dunes} use terrestrial radar and topography to model dune backscatter and examine
how those models might apply to Titan. \citet{cartwright2017alluvial} compare terrestrial and Titan
radar observations to test fan formation mechanisms. Such interpretations require observations
beyond an image classification score. CETUS predicts broad map units from one display mosaic; its
disagreements identify images that warrant closer interpretation. Determining a scattering mechanism
or surface material requires additional physical evidence.

\paragraph{Evaluating transferred representations:}
\citet{kornblith2019transfer} compare classifier training on existing image features with fine-tuning across image classification datasets. \citet{kumar2022distortion} identify settings where full fine-tuning performs worse under distribution shift than a linear classifier trained on the original features. These findings motivate evaluating the original representations before training them further. CETUS trains only part of each encoder under one set of settings. Testing the explanation based on feature distortion proposed by Kumar and colleagues would require a separate analysis.

\paragraph{Untrained representations:}
\citet{saxe2011random} showed that random weights in convolutional square-pooling networks could produce useful features for a supervised image classifier. Their experiments also distinguished the contributions of architecture and weight learning and examined variation across random initializations. Those results concern convolutional networks. They motivate evaluating random features as a baseline, while the behavior of CETUS's Untrained ViT must be assessed from its own predictions.

\subsection{Model comparisons}

Encoder rankings depend on classifier fitting in the geographic evaluation. CROMA has the highest
mean F1 with logistic regression, while DINOv2 leads with SGD and after training on Titan. CROMA's
logistic regression result is already similar to its mean F1 after encoder training. DINOv2 improves
over both classifiers in every fold, whereas DOFA declines against both. An encoder's ranking is
therefore conditional on the classifier fitting settings, the encoder weights allowed to change and
the evaluation images. Encoder training also changes optimization, regularization, batch sizes,
numerical precision, training duration and the epoch used for testing. The results do not separate
those contributions, and uncertainty about DOFA's pretraining inputs further limits interpretation
of its decline. The scores measure agreement with the map under the evaluated configurations.
Geological accuracy and usefulness in practice require independent evaluation.

A direct test of pretraining would compare learned weights with multiple random encoder
initializations while holding architecture, input processing and classifier fitting constant.
Training a small convolutional neural network (CNN) on the current Titan folds would add a compact
supervised comparison. Both experiments remain future work. The existing CNN results in
Appendix~\ref{app:original_dataset} use the original dataset and geographic divisions, so they
cannot supply that comparison on the current folds.

The MMD and t-SNE analyses describe feature distributions in the original dataset. MMD depends on
the encoder, kernel and sampling scheme; t-SNE displays a projection of the sampled features.
Classification on separate Titan test images is needed to assess predictive usefulness. CORAL
adjusts Earth feature statistics using Titan training features, and its test recalls quantify the
predictive effect. The smaller MMD after CORAL uses the fitting samples and a newly estimated
bandwidth, leaving a reduction on independent data under a common kernel untested. These comparisons
also leave the physical causes of domain differences unresolved.

\subsection{Terrain interpretation and data limitations}
Terrain predictions follow the categories and mapping scale of the reference map. Better agreement
for Labyrinths alone leaves open whether a model recognizes dissection. A Craters prediction can
refer to a rim, ejecta or fill within a large image area; identifying distinct impact structures is
a separate task. Disagreements can arise from mixed terrain, simplified map edges, information
available to the mapping team outside the SAR image, or model error. The scores do not distinguish
among these possible causes. Examining the images alongside acquisition metadata and an independent
geological interpretation would help distinguish them.

Using the same mosaic makes the inputs comparable across methods. Within that mosaic, SAR and HiSAR
modes provide different spatial information, effective resolution varies geographically, and image
rendering affects brightness and texture. A follow-up analysis could compare errors for the same map
classes across verified observing conditions, with the comparison specified before its results are
examined. It could test whether acquisition contributes to the observed disagreements between image
predictions and map labels.

\begin{samepage}
Nearby SGD test errors often covary within a map class and geographic region.
Appendix~\ref{app:errors_by_distance} gives the counts and assumptions behind those correlations.
Quantifying any score inflation or choosing an adequate separation distance requires further
experiments.

\end{samepage}

The expanded catalog and geographic divisions both differ from the original evaluation, so
differences in scores combine changes in longitude handling, regional grouping and spatial
separation. The five sectors sample the available display mosaic unevenly in both observing
conditions and terrain classes. Class mixture, spatial dependence and the small number of regions
limit generalization beyond those images. Independent expert annotations would help assess
disagreement with the map. Agreement with those annotations and reproduction of the map's labels
remain distinct evaluations.

An expert label review should distinguish the SAR images available to the classifier from the
complementary observations used by the mapping team. It should include mixed images, classes with
few examples and multiple regions. Agreement among experts would measure consistency under the
stated viewing conditions; it would not set a universal ceiling on a classifier's agreement with one
map. Assessing geological accuracy requires observations with an appropriate independent basis.
Further model training would not supply that independent evidence.

The Earth experiments also need better source labels. Their inputs include incorrectly named
locations, offshore images with uncertain terrain assignments and overlapping training and
validation areas. Equalizing numerical brightness ranges leaves the underlying radiometric
differences unresolved. The current Titan experiments keep the image product and class definitions
common to all models, allowing comparison of geographic map prediction and sensitivity to classifier
fitting. Separating radar physics, image rendering and representation learning as causes of errors
requires the additional physical and experimental evidence described above.

\section{Conclusion}
In the evaluated configurations, terrestrial representations contain useful information for
predicting the terrain classes in Titan's geomorphological map. Classifier fitting affects how that
information translates into predictions: the leading encoder changes when feature scaling,
optimization and regularization change together. Training the last two DINOv2 layers improves
regional macro F1 over both classifier baselines and improves Dunes, Hummocky and Labyrinths F1 over
logistic regression in every fold. CROMA's small overall change conceals opposing changes among
classes, while DOFA declines under its evaluated input and training settings. Craters remains poorly
predicted, and its test tiles are concentrated in a small number of map components. The CETUS
predictions allow these comparisons to be reproduced and identify disagreements for geological
inspection. Independent evaluation is still needed before using the models for mapping or image
search.

\clearpage
\renewcommand{\bibfont}{\raggedright}
\setlength{\bibsep}{4pt}
\bibliographystyle{unsrtnat}
\bibliography{references}
\clearpage

\appendix

\section{Further results from the geographic evaluation}

The geographic evaluation compares predictions with an expert map and checks that saved models
reproduce those predictions. This appendix gives the training totals, additional class results and
spatial analysis of the errors. The complete class metrics and sample counts for training the final
two transformer blocks appear in the main Results. Appendix~\ref{app:original_dataset} describes the
original dataset and the additional Earth and Venus analyses associated with
Section~\ref{sec:earth_venus_comparison}.

\subsection{Reproducing predictions}

Saved estimators reproduce every predicted class in the common logistic regression comparison, with
probabilities agreeing to an absolute tolerance of $10^{-12}$. Independent
calculations also reproduce the original SGD and nearest-neighbor predictions, repeat normalization
from raw pixels, and verify the confusion matrices and metrics for training the final two
transformer blocks. These calculations reuse saved features, without repeating pretrained inference.

CETUS's reproduction documentation links the catalog, geographic splits, predictions, metrics for
each fold and counts by map polygon part. It also identifies the common logistic regression and
classical-feature results, experimental specifications and separate Visual and Infrared Mapping
Spectrometer (VIMS) display metadata with corrected coordinates. The files for the original dataset
retain their earlier contents.

The documentation also distinguishes the software used for the original experiments from the public
software used to reproduce them. It retains all fold values, warnings, fits that reached their
iteration limit and sample counts, with the complete training histories and instructions for using
saved predictions or starting new training.

\subsection{Training the final two transformer blocks}
\label{app:encoder_training}
Section~\ref{sec:experiment_design} specifies training and metric averaging. The saved models
correspond to the evaluation epochs, while the computation totals include training after those
epochs.

All 75 saved models chosen by validation reproduced the saved predictions. The epochs used for
evaluation range from 1 to 14 for DINOv2, 1 to 21 for DOFA and 1 to 13 for CROMA; every fit
continued through epoch 50. Fold 1 contains a single validation tile labeled Craters, so that tile
determines Craters recall in the validation metric. Agreement between repeated predictions verifies
the calculation. Geological interpretation still depends on the map labels, class mixture, spatial
dependence and test counts discussed in the main text. Table~\ref{tab:training_completion} gives the
complete training totals. Tables~\ref{tab:encoder_training_scores}
and~\ref{tab:encoder_training_classes} give the macro comparison and class ratios.

\begin{table}[ht]
\centering
\caption{Training totals when fitting the final two transformer blocks. Epochs start at one. The
ranges identify epochs used for evaluation; the computation totals cover all 50 epochs in 25 fits
per encoder. Example presentations count each use of an image, including repetitions across epochs
and fits. Successful updates and updates skipped by automatic mixed precision (AMP) sum to attempted
updates. Comparing computational costs with experiments that keep encoder weights unchanged requires
the corresponding totals for those experiments.}
\label{tab:training_completion}
\small
\setlength{\tabcolsep}{4pt}
\begin{tabular}{lrrrrr}
\toprule
Encoder & \shortstack{Epoch range\\used for evaluation} & \shortstack{Attempted\\updates} & \shortstack{Successful\\updates} & \shortstack{AMP\\skips} & \shortstack{Example\\presentations} \\
\midrule
DINOv2 & 1--14 & 527{,}750 & 527{,}475 & 275 & 16{,}871{,}750 \\
DOFA & 1--21 & 527{,}750 & 527{,}453 & 297 & 16{,}871{,}750 \\
CROMA & 1--13 & 527{,}750 & 527{,}487 & 263 & 16{,}871{,}750 \\
\bottomrule
\end{tabular}
\end{table}

\subsection{Class predictions with SGD}

The random encoder gives macro recall close to one-sixth under SGD. When all six classes occur in a
test set, predicting any one of those classes for every tile gives exactly one-sixth macro recall.
This equality follows from the metric and does not define a chance null distribution. With the
classifiers fitted here, the pretrained features give better agreement with the map than this
particular random encoder.

In fold 2, each of DOFA's five SGD classifiers correctly identifies the only Craters tile but
assigns that class to 164 to 171 of the 4{,}497 test tiles. Across folds, mean Craters precision is
1.99\%, recall 40.85\% and F1 3.57\%. The additional Craters predictions disagree with the most
common map class in those tiles; some tiles may nevertheless contain mapped crater terrain. A
separate geometric check found that each sampled Craters polygon part belongs to only one of
training, validation or testing in a given fold. Several parts may describe the same physical
crater, so part counts cannot establish the number of independent craters.

\begin{table}[ht]
\centering
\caption{Precision, recall and F1 by map class for SGD classifiers on the enlarged catalog, in
percent. Entries are precision / recall / F1. Each ratio is calculated for each classifier and fold,
then averaged over classifier seeds in the fold and equally over folds.
Figure~\ref{fig:titan_geography} gives the class counts. Bold indicates the highest mean F1 across
encoders for each class.}
\small
\begin{tabular}{lrrr}
\toprule
Class & DINOv2 & DOFA & CROMA \\
\midrule
Plains & 85.93 / 67.60 / \textbf{75.54} & 86.57 / 42.92 / 56.61 & 88.74 / 55.25 / 67.78 \\
Dunes & 53.86 / 69.11 / \textbf{60.21} & 48.77 / 59.99 / 52.67 & 48.03 / 59.81 / 51.78 \\
Hummocky & 39.06 / 59.61 / 46.03 & 43.49 / 47.75 / 39.30 & 48.22 / 58.61 / \textbf{49.37} \\
Labyrinths & 24.78 / 45.62 / \textbf{29.85} & 6.70 / 48.51 / 11.45 & 12.12 / 57.02 / 19.12 \\
Lakes & 9.94 / 20.01 / 11.29 & 7.90 / 31.51 / 10.67 & 16.03 / 42.12 / \textbf{19.97} \\
Craters & 2.56 / 6.24 / 3.33 & 1.99 / 40.85 / \textbf{3.57} & 1.98 / 17.90 / 3.53 \\
\bottomrule
\end{tabular}
\end{table}

\begin{figure}[ht]
\centering
\includegraphics[width=\linewidth]{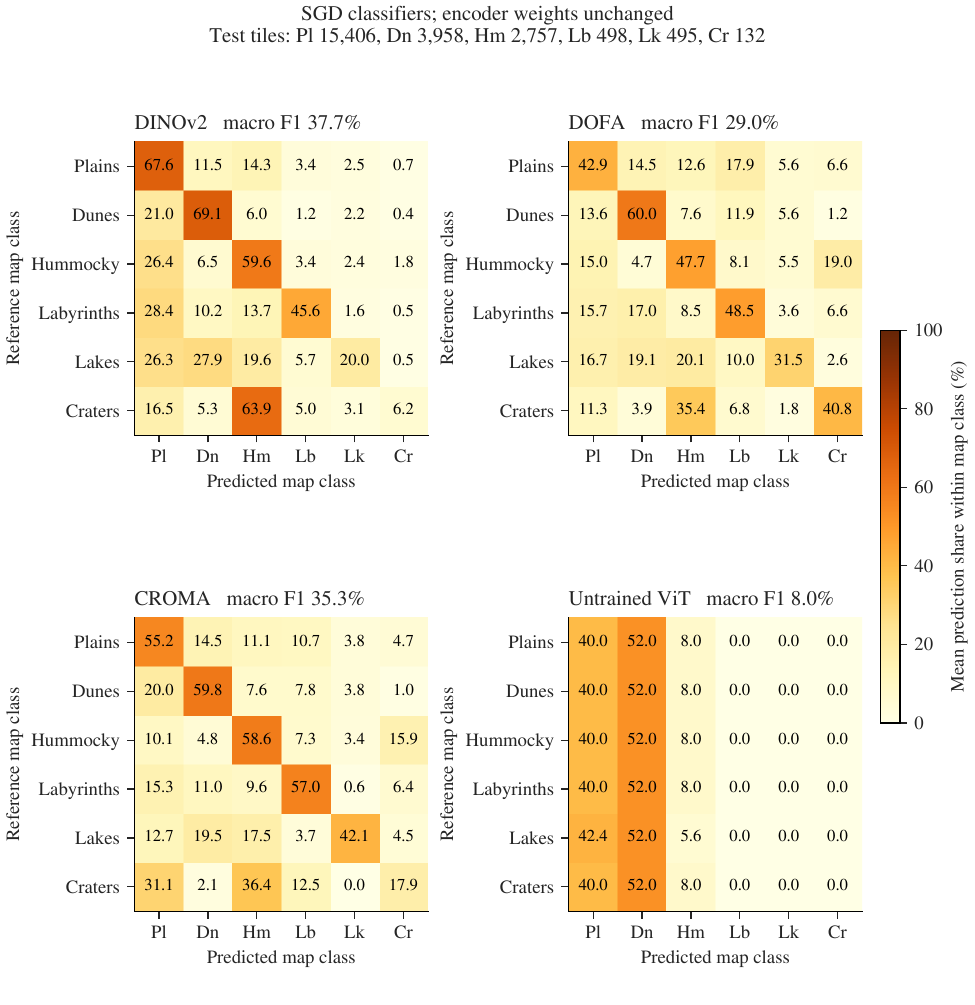}
\caption{SGD classification in the current geographic evaluation. Rows give the map class; columns
give the predicted class. Percentages first average classifier seeds in each fold, then give all
folds equal weight. The count line includes each distinct test tile once across folds, irrespective
of the number of classifier seeds. Diagonal values are mean class recall. Precision and F1 also
depend on predictions for other map classes; their primary estimates average ratios calculated
before combining folds. Craters counts range from one to 61 tiles per fold. The labels describe
broad mapped units.}
\label{fig:sgd_confusion}
\end{figure}

\subsection{How errors vary with distance}
\label{app:errors_by_distance}

The spatial analysis groups the original SGD and nearest-neighbor test predictions into 30
combinations of fold and map class. At each tile, the SGD error is the mean of five binary
classifier errors. In each group, unordered pairs of tile centers are enumerated using great-circle
distance on a sphere of radius 2{,}574.7 km. Distance bins are $(0,45]$,
$(45,90]$, $(90,180]$, $(180,360]$ and
$(360,720]$ km. Symmetric Pearson correlation is calculated by entering each pair in
both orders and correlating the two error values. Pairs receive equal weight, so a tile contributes
once for each pair in which it occurs. A coefficient requires at least 20 tiles in the group, 20
participating tiles and variance above $10^{-12}$ among the values being
correlated. Otherwise the result includes the reason that the coefficient is undefined. This
analysis uses the original classifiers; the primary logistic regression fits are evaluated
separately.

The analysis also regresses the SGD error fraction or the nearest-neighbor binary error on
coordinates on the unit sphere, the fraction of the tile occupied by its modal map class, and the
valid SAR fraction, with an intercept and a separate fit for each group. The same correlations are
then calculated from the residuals. Removing a fitted trend can create apparent spatial structure.
For CROMA's nearest-neighbor classifier in fold 4, all 76 Lakes tiles participating in the first
distance bin are misclassified, giving zero variance before regression. Only one of the 102 tiles in
that group is classified correctly. After regression, the residual correlation is 0.99693 across 52
pairs. The high coefficient thus characterizes the residual errors and offers no evidence for a
physical mechanism. CETUS's reproduction documentation identifies the complete unadjusted and
residual coefficients, sample counts, undefined cases and independent numerical reconstruction.

These correlations describe whether nearby test tiles tend to be misclassified together. Estimating
leakage between training and testing, inflation of accuracy, or effective sample size would require
further analysis. Section~\ref{sec:training_geography} gives the separate experiment, specified in
advance, that changes which nearby tiles enter training.

\section{Transfer and feature distributions in the original dataset}
\label{app:original_dataset}

The original catalog supplies the Earth transfer comparisons and unlabeled Venus measurements in
Section~\ref{sec:earth_venus_comparison}. It uses different image samples, geographic divisions and
fitting settings from the current Titan evaluation. The transfer and distribution results below
therefore refer to the original populations.

\subsection{Images and geographic divisions}

\begin{figure*}[htbp]
\centering
\includegraphics[width=\textwidth]{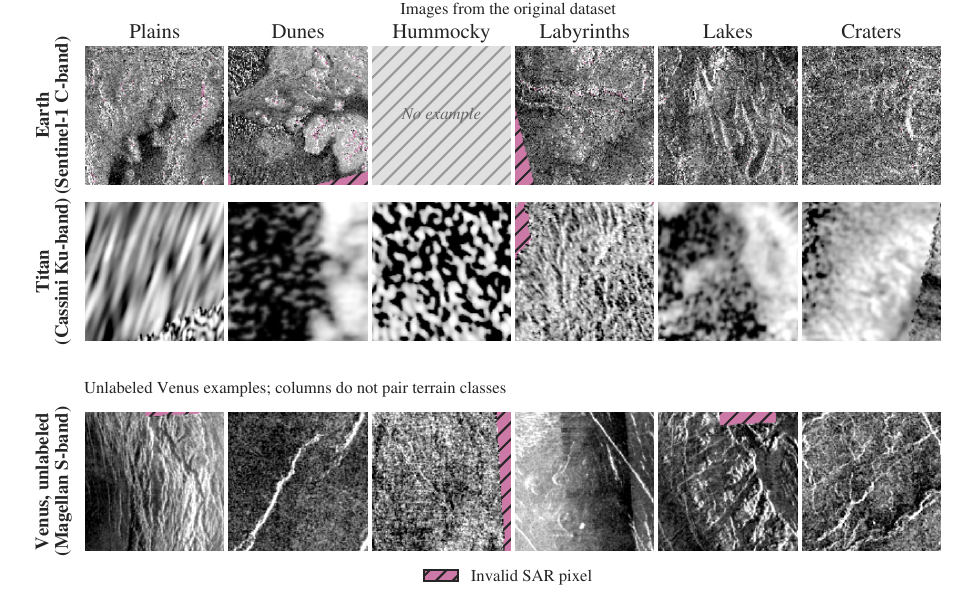}
\caption{Examples from the original Earth, Titan and Venus collections. Columns show Earth site
labels, whose landform assignments still need review, and Titan classes from the expert map. Earth
contains no Hummocky examples. Venus images have no class labels or implied class correspondence
across columns. Each image spans a nominal \SI{45}{\kilo\meter} by \SI{45}{\kilo\meter}. Earth
linear $\sigma^0$ is shown as $10\log_{10}(\sigma^0)$ in dB; Titan and Venus
use display digital numbers. Each display stretches the 2nd to 98th percentiles of valid pixels, so
brightness is relative to that image. Magenta and diagonal hatching identify invalid SAR pixels,
including Earth pixels filled during preprocessing. The examples illustrate image appearance;
geological equivalence requires independent evidence.}
\label{fig:radar_examples}
\end{figure*}

\subsubsection{Choosing the image examples}
The examples retain the original image choices. The statistics used to rank them were calculated
from finite, strictly positive pixel values in the stored arrays before conversion for display. For
the Earth display, missing-pixel fills are now masked using the catalog validity counts and excluded
from the percentile stretch; the original image arrays and selection statistics are unchanged. For
Earth and Titan, the score favors larger standard deviation and penalizes variation among row means.
Plains examples receive an additional preference for standard deviations near the median among
eligible images. For Venus, a contrast filter first removes unsuitable examples; the remaining image
with the highest score, including the same row-variation penalty, supplies the first example. Each
statistic is scaled linearly to the unit interval over eligible images. Five more images are then
chosen successively, maximizing the minimum distance to those already chosen in mean, standard
deviation and skewness.

\subsubsection{Cassini SAR imagery}

The Titan images in Figure~\ref{fig:radar_examples} use the USGS HiSAR mosaic through T104 described
in Section~\ref{sec:titan_data} \citep{usgs2015titan,elachi2004cassini}. Its 8 bit logarithmic
display values retain image appearance but do not provide calibrated absolute
$\sigma^0$. Physical comparisons of radar return would require radiometric products
such as the floating point backscatter images in Cassini Basic Image Data Records
(BIDR)~\citep{wall2019radarguide}.

\paragraph{Image area and valid pixels:}
Each Titan and Venus tile is sampled in a local azimuthal equidistant projection whose origin is the
tile center. Earth tiles are windows in the local UTM output of SNAP processing. Every tile
represents \SI{45}{\kilo\meter} by \SI{45}{\kilo\meter} with $128 \times 128$ pixels and
a spacing of \SI{351.5625}{\meter}/pixel. Local metric sampling avoids the change in physical width
with latitude that would arise from equal pixel windows in the global equirectangular raster. At
least 80\% of output pixels must satisfy the validity rule for the corresponding image product.
The original catalogs contain 11{,}371 labeled Titan tiles: 11{,}306 for cross-validation and 65
reserved near Selk. They also contain 398 labeled Earth tiles from 13 analog sites and 2{,}945
unlabeled Venus tiles. Titan counts range from 7{,}458 Plains tiles to 25 Craters tiles. Macro
recall gives each class equal weight, making the very small Craters sample particularly
consequential for interpretation.

The USGS mosaic ends at T104, earlier than the observations used to calculate the mission totals of
approximately 46\% surface coverage by full-resolution SAR and 74\% by SAR with
HiSAR~\citep{wall2019radarguide}. This evaluation covers only the image areas retained from the T104
mosaic.

\paragraph{Image normalization:}
For each Titan training fold, the 1st and 99th percentiles of a limited sample of display DN define
the linear normalization. The Titan transformation uses the display values directly, without an
additional logarithm or dB floor. Its output remains display data; recovering calibrated
$\sigma^0$ would require a different input product. For Earth, calibrated
Sentinel-1 $\sigma^0$ is converted to dB and scaled using training statistics.
Venus FMAP values are normalized in their encoded DN space. Each transformation prepares model
inputs in the units available for that product, retaining the differences in their radiometric
meaning.

\subsubsection{Map labels}

Both Titan evaluations use the map classes and modal voting rule described in
Section~\ref{sec:titan_data}. Using the same classes permits comparison of the two evaluations
against the map. Extending those labels to Earth also requires evidence about the surface materials
and class distributions represented at terrestrial sites.

\paragraph{Recovering omitted western longitudes:}
The original sampler interpreted the map raster's 0 to 360 degree longitude convention incorrectly.
Reprojection returned missing data for valid western labels, and catalog generation then discarded
the corresponding tiles. Explicit longitude wrapping and a smaller warp approximation tolerance
recover 12{,}009 tiles, including 115 Craters. The new catalog preserves all 11{,}371 original image
arrays and centers while changing 15 of their modal labels and 1{,}727 class fractions. Independent
inverse projections and direct indexing of the map raster agree with all 168{,}501{,}248 samples at
the 41{,}138 possible tile centers. The resulting catalog contains 23{,}380 tiles with at least 80\%
valid SAR coverage and mapped terrain, including 140 Craters. These checks verify sampling of the
source map. Geological accuracy and classification performance require separate evidence. The main
text evaluates the enlarged catalog using geographic sectors with separation zones; CETUS's
reproduction documentation links the label and catalog checks.

Independent repeat annotations and a numerical estimate of map error remain unavailable. A
classifier can reproduce an error in the expert map, so agreement with that map may exceed agreement
with independently established geology.

\subsubsection{VIMS and ISS display metadata}

The catalog also contains display values from the combined VIMS and ISS mosaic of
\citet{seignovert2019mosaic}; processing of its VIMS component is described by
\citet{lemouelic2019vims}. ISS is the Imaging Science Subsystem. The original sampler supplied
longitude and latitude in degrees to a raster transform expressed in metres on a Titan sphere of
radius 2{,}575{,}000 m. It also used coordinates of pixel corners as array indices. Consequently,
all 23{,}380 auxiliary triplets in the enlarged catalog come from just six pixels near the map
origin. A separate CETUS file uses the correct coordinate units and pixel centers and passes
independent projection and interpolation checks. The previously distributed catalogs retain their
original values.
These channels describe a rendered composite. Quantitative spectra, radiances and band ratios would
require calibrated VIMS products, and describing an entire image area would require an explicit rule
for combining measurements across that area. CETUS restricts these triplets to display metadata.
Terrain classification, comparisons of feature distributions and Selk scores exclude them.
Quantitative use of the imaging systems would require separate validation.

\subsubsection{Earth source sites}

The Earth sites were chosen as morphological analogs for \textbf{Plains}, \textbf{Dunes},
\textbf{Labyrinths}, \textbf{Lakes} and \textbf{Craters}. The Earth set contains no Hummocky
examples, so transfer metrics omit that class. The site labels describe intended landform analogies.
Equivalence in material, radiometry or formation process remains unestablished.
Several assignments need review at the scale of individual images. Sixteen training tiles labeled
Labyrinths come from an Egyptian region incorrectly named Wadi Rum in the catalog. The Sentinel-1
image extents establish the location error; changing the name alone would leave the terrain
assignment unverified. The full-resolution GSHHG shoreline map shows no mapped land in 42 of the 77
Arctic tundra training areas, and marine regions occupy more than half of 65 such areas. One
Icelandic Plains tile is mostly ocean in both global and national shoreline maps. These fractions
use the nominal image areas without the SAR validity mask. The shoreline maps also leave sea ice and
open water indistinguishable at the acquisition dates. A review of the actual image areas is
therefore needed before using their uniform site labels as terrestrial Plains examples. The internal
geometry analysis is identified in CETUS's reproduction documentation.\footnote{Shoreline
documentation: \url{https://www.soest.hawaii.edu/pwessel/gshhg/}.}
Appendix~\ref{app:dataset} lists the sites and the intended morphological rationale. Earlier SAR
studies illustrate how terrestrial landforms can inform comparative geomorphological
interpretation~\citep{paillou2014dunes,cartwright2017alluvial,miller2021fluvial}.

ESA SNAP processing of Sentinel-1 GRD applies a precise orbit file, removes thermal noise,
calibrates VV to linear $\sigma^0$, and applies Range-Doppler terrain correction
with the Copernicus \SI{30}{\meter} DEM. The terrain-corrected output is then sampled at
\SI{351.5625}{\meter}/pixel over each \SI{45}{\kilo\meter} image area. GRD processing already
includes detection and averaging of radar looks, called multilooking. The SNAP sequence adds no
further averaging before coarse resampling. The source spacing of about \SI{10}{\meter} describes
the sampling grid, rather than the effective spatial resolution.
Pixels that are invalid or marked as layover or shadow by SNAP enter the encoder at the
$-40$ dB floor. They are excluded from each tile's dB statistics and
valid-pixel fraction. For every retained tile, the median local incidence angle and the fraction
marked by that mask are also retained.

Earth validation uses whole Sentinel-1 products as groups. Slices from one acquisition and repeat
views of the same terrain can nevertheless enter both training and validation. Reconstructing the
nominal image areas shows that 21 of 46 validation tiles intersect training areas; two Mistastin
pairs overlap by about 99.97\%. Grouping by product therefore permits geographic overlap between
training and validation.
For transfer, a reproducible rule reserves one product per Earth class for validation. The 46
validation tiles include the only product from one site, leaving 352 training tiles across 12 sites.
The resulting classifier is evaluated on the five Titan classes represented in Earth training.

\subsubsection{Geographic folds in the original evaluation}

The original Titan evaluation uses five-fold spatial block cross-validation. Here the blocks are
metric regions approximately \SI{250}{\kilo\meter} across, with longitude divisions adjusted for
latitude. A reproducible greedy assignment keeps each region together while balancing class counts
and total image counts across folds. Each run tests one fold, uses the next for validation and
trains on the remaining folds. Neighboring regions may have different roles, with no additional
separation zone between them. Nearby images can therefore remain correlated between training,
validation and testing. Training sets also overlap across runs, so independence of the five
test-fold estimates has not been established.
The 17 Craters test tiles receive their labels from four polygon parts in the source map. Parts 14
and 18 contribute both training and test tiles in individual runs. These indices describe map
geometry and cannot be equated with named or independent physical craters. They establish that
training and testing include portions of the same mapped terrain, with an unmeasured effect on
performance.
Using the benchmark's spherical Titan radius of \SI{2574.7}{\kilo\meter}, 1{,}963 of the 11{,}306
distinct test centers lie less than \SI{45}{\kilo\meter} from a training center in the corresponding
run; only 79 lie less than \SI{44.9}{\kilo\meter} away. Many distances are close to the nominal
\SI{45}{\kilo\meter} grid spacing, making counts sensitive to the radius and strict cutoff. These
measurements describe proximity of centers. Image-area overlap, similarity of radar texture and any
inflation of performance require separate measurements.

The Selk region is removed before assigning the other regions to folds and is excluded from every
training, validation and test set. Results are summarized over the five test folds. Repeated SGD
training runs describe variation from initialization and minibatch order within each fold; they do
not create additional independently sampled regions.

\subsection{Encoders and classifiers for the original dataset}

\subsubsection{Encoder configurations}
\label{sec:model_descriptions}

The four encoders use ViT-B~\citep{dosovitskiy2021vit} and return 768-dimensional feature vectors.
The comparison evaluates complete configurations: pretraining, patch sizes, input transformations,
conditioning and architecture details vary together.

DINOv2 and CROMA use the image dimensions, channel duplication and feature averaging specified in
Section~\ref{sec:experiment_design}. DINOv2 was pretrained on 142M natural images
\citep{oquab2024dinov2}. CROMA averages its patch encodings and passes that mean through the learned
feedforward transformation for SAR \citep{fuller2023croma}.

\paragraph{DOFA ViT-B/16:} A hypernetwork generates patch embedding weights from a numeric band identifier~\citep{xiong2024dofa}. Titan normalization and feature averaging follow Section~\ref{sec:experiment_design}. The original Earth features use a different band identifier from the repository example.
The DOFA repository example supplies $5.405$ for Sentinel-1; CETUS extends this
frequency convention to Cassini Ku-band with $13.78$. Optical bands use
wavelength units. Inspection of the extraction records and arrays establishes that the original
five-fold classification used $13.78$ for both Earth and Titan. The earlier DOFA
paper gives 3.75 for Sentinel-1, leaving the identifiers used in pretraining this checkpoint
unresolved \citep{dofapreprintv2,dofareadme2026}. Arrays generated separately for preprocessing
comparisons and Venus need their own verification. Section~\ref{sec:earth_preprocessing} describes
later experiments using Earth 5.405 and Titan 13.78, together with other changes to the experimental
design. Measuring the effect of the identifier alone would require an otherwise identical
comparison.

\paragraph{Untrained ViT in the original dataset:}
The original Untrained ViT configuration specifies a ViT-B/14, a single-channel patch stem and seed
42. Its encoder remains untrained while the classifier learns the labels supplied by the
corresponding training domain. Feature standard deviations are approximately
$10^{-4}$ per dimension, and the classifiers perform near the class-frequency
benchmarks. Because this result comes from one random state, it gives limited evidence about random
features in general. Section~\ref{sec:feature_distributions} separates the valid repeated MMD
estimate from the numerical failure in the original permutation calculation.
The code specifies seed 42, but the architecture chosen by its fallback branch and the exact initial
weights remain unverified. Five classifier seeds therefore cannot be interpreted as five
independently initialized encoders.

\subsubsection{Fitting classifiers and calculating metrics}

With encoder weights unchanged, each model returns 768 features through the averaging or
classification-token representation described in the main methods. All 168 tensors loaded from the
original DOFA checkpoint were verified exactly. The normalization constants supplied when the model
is constructed and the checkpoint normalization parameters that remain unused are the same as in the
current evaluation; their separate effects on performance remain unknown.
The linear classifier is fitted for 100 epochs with inverse-frequency class weights, and the final
epoch is evaluated on the test images. The SGD implementation does not use the reserved validation
fold for model selection. Its main metric is macro recall, the unweighted mean of recall over
classes present in the evaluation set, equivalent here to balanced accuracy. Cohen's
$\kappa$ compares agreement with that expected from the marginal prediction and
label frequencies. Its dependence on class frequencies differs from that of macro recall, and
per-class precision supplies additional information. Both metrics are calculated for five training
seeds in each geographic fold. A cosine $k$-nearest-neighbor classifier
supplies a comparison without fitting a linear decision rule. Appendix~\ref{app:implementation}
lists the image inputs and weight sources.

For the Titan-only experiment, classifiers are fitted to Titan training data and predict the
reserved Titan test fold. For transfer, they fit the grouped Earth training data and are evaluated
on the five Titan test classes represented in Earth training. The number of Hummocky images omitted
from that evaluation is also reported. Cross-validation statistics use one estimate per fold.
Classifier training seeds and repeated MMD samples describe variation within folds; the number of
geographic test units remains five.

Additional comparisons on the original dataset include prediction of the majority class, random
prediction using class frequencies, raw intensity moments, intensity histograms, gray-level
co-occurrence matrix (GLCM), local binary pattern (LBP), Gabor, histogram of oriented gradients
(HOG) and wavelet texture features. Logistic regression on combined classical features and a small
supervised CNN are also evaluated on that dataset. The class-frequency comparisons use no image
information. Classical features and the CNN assess prediction from images, while partial encoder
training and correlation alignment (CORAL) test additional uses of Titan data. The original CNN
results do not evaluate the current geographic folds.
CORAL uses unlabeled Titan training features for unsupervised domain adaptation and evaluates
classification on separate Titan test regions. The metadata call this procedure transductive, but
the transformation uses only training features. MMD calculated from the fitting features describes
agreement on those training samples.

\subsection{Comparing feature distributions}

\subsubsection{Estimator and sampling details}
\label{sec:metrics}

Section~\ref{sec:venus_comparison} defines MMD and the comparisons between Earth, Titan and Venus
image collections. The primary estimate uses the median heuristic for kernel bandwidth and averages
at least 100 reproducible subsamples with equal $N$ from each distribution.
Estimator sensitivity is examined with the unbiased U-statistic, a permutation test using one
bandwidth throughout, omission of each Earth site in turn, and resampling of Titan geographic
regions with replacement. Section~\ref{sec:mmd_sensitivity} gives the results, with complete values
in CETUS.
The site omissions cover the 12 sites in Earth training; validation contains the only product from
the thirteenth site. Titan resampling uses the same geographic regions that define the folds.
Dependence between neighboring regions remains unmeasured.

Proxy $\mathcal{A}$-distance~\citep{bendavid2010theory} provides a second measure of
how readily a classifier distinguishes Earth from Titan features. The original calculation used
different numbers of images from the two domains. The estimates for all twenty combinations of model
and fold were subsequently recomputed from the verified original training features, with equal
domain counts, a stratified half split, standardization fitted on the training half used for the
domain classifier, hinge-loss SGD and balanced test error. Both calculations use the same data
population but differ in sampling and classifier fitting. Table~\ref{tab:feature_distributions}
reports the original estimates; the later estimates are given below. Both measure domain
discrimination; assigning a physical cause requires other evidence. Distances between class
centroids are also descriptive and use only classes labeled in both domains.
\FloatBarrier

Attributing these differences to radar scattering would require compatible calibrated products and
measurements that account for incidence, polarization, spatial resolution, roughness, dielectric
properties and preprocessing. HiSAR and FMAP supply display values here. The rendering experiments
test model sensitivity to those display inputs. Recovering calibrated planetary backscatter would
require suitable radiometric products.

\subsection{Further results on the original dataset}

\subsubsection{Training and averaging across folds}

Feature extraction uses batches of 64 images. Bilinear interpolation resizes each
$128 \times 128$ tile to $224 \times 224$ pixels for DINOv2, DOFA and the
Untrained ViT, or $120 \times 120$ pixels for CROMA. Linear classifiers train in batches
of 256 for 100 epochs, with five training seeds per fold.
Averaging estimates from the five training seeds gives one estimate per fold. The reported mean and
sample standard deviation are then calculated across the five folds. The 95\%
Student-$t$ intervals and paired tests supplied with the results assume
independent fold estimates. Overlapping training data and possible correlation between neighboring
test regions make that assumption uncertain, so these calculations are interpreted as
exploratory.\footnote{Bengio and Grandvalet, ``No Unbiased Estimator of the Variance of K-Fold
Cross-Validation,'' JMLR 5, 2004: \url{https://jmlr.org/papers/v5/grandvalet04a.html}.} Differences
use unrounded values and may differ by up to 0.1 from subtracting rounded table entries. The
$k$-NN classifier uses $k=20$ and gives the same result
each time for a given fold.
The distribution comparisons use 100 reproducible MMD subsamples of equal size per fold, the
exploratory fold comparisons and the sensitivity calculations in Section~\ref{sec:metrics}. All
results here use the original population, including the omission of western Titan tiles.

\subsubsection{Training and testing on Titan}

Table~\ref{tab:within_domain} gives macro recall for classifiers trained and tested on separate
Titan regions.
\begin{table}[t]
\centering
\caption{Training and testing on Titan in the original catalog. Values give means and sample
standard deviations over five geographic folds. For linear classifiers, each metric is first
averaged over the five training seeds in each fold; $k$-NN gives one
reproducible result per fold. Among linear classifiers, CROMA has the highest six-class macro
recall, 66.1\%, and DINOv2 the highest $\kappa$, 0.502. Untrained ViT uses an
untrained encoder whose original architecture remains incompletely verified, as described in
Section~\ref{sec:model_descriptions}. Bold indicates the highest mean in each metric column for its
classifier.}
\label{tab:within_domain}
\begin{tabular}{lcccc}
\toprule
\textbf{Model} & \multicolumn{2}{c}{\textbf{Linear classifier}} & \multicolumn{2}{c}{\textbf{$k$-NN}} \\
\cmidrule(lr){2-3} \cmidrule(lr){4-5}
 & Macro recall & $\kappa$ & Macro recall & $\kappa$ \\
\midrule
DINOv2 ViT-B/14       & 56.0 $\pm$ 2.7 & \textbf{0.502} $\pm$ 0.039 & 32.4 $\pm$ 1.4 & 0.388 $\pm$ 0.016 \\
DOFA ViT-B/16          & 58.1 $\pm$ 8.0 & 0.296 $\pm$ 0.030 & 33.5 $\pm$ 0.8 & 0.400 $\pm$ 0.020 \\
CROMA ViT-B/8          & \textbf{66.1} $\pm$ 2.4 & 0.432 $\pm$ 0.038 & \textbf{38.7} $\pm$ 0.7 & \textbf{0.440} $\pm$ 0.034 \\
Untrained ViT-B/14   & 17.0 $\pm$ 0.7 & 0.003 $\pm$ 0.006 & 19.0 $\pm$ 0.9 & 0.022 $\pm$ 0.010 \\
\bottomrule
\end{tabular}
\end{table}

Table~\ref{tab:baselines} compares classification from simple image measurements with predictors
based only on class frequencies and a supervised CNN. The supervised CNN reaches 64.4\% on Titan and
19.0\% when trained on Earth. Raw intensity statistics give the highest transfer mean among these
methods, 28.1\%. CROMA reaches 32.3\%, exceeding it by 4.2 percentage points. With five folds, the
exploratory paired test gives $p=0.217$, leaving evidence for a transfer advantage
inconclusive.
\begin{table}[t]
\centering
\caption{Additional classification comparisons on the original dataset. Macro recall is in percent,
with mean $\pm$ sample standard deviation across five geographic folds after
averaging seeds in each fold. Classical-feature classifiers give identical results across seeds;
only class-prior random predictions and CNN fits vary with seed. Transfer excludes Hummocky.
Classical classifiers fit the five Earth classes, and the CNN's Hummocky logit is masked before
softmax. All CNN predictions before masking already belong to a represented class, so masking leaves
the test predictions unchanged. Bold indicates the highest mean recall in each training-domain
column among the methods shown.}
\label{tab:baselines}
\begin{tabular}{lcc}
\toprule
\textbf{Comparison} & \textbf{Titan$\to$Titan} & \textbf{Earth$\to$Titan} \\
\midrule
Majority class & 16.7 $\pm$ 0.0 & 20.0 $\pm$ 0.0 \\
Class-prior random & 16.7 $\pm$ 0.2 & 20.0 $\pm$ 0.9 \\
Raw intensity statistics & 54.8 $\pm$ 7.3 & \textbf{28.1} $\pm$ 4.9 \\
Intensity histogram & 42.8 $\pm$ 4.8 & 19.7 $\pm$ 1.0 \\
Texture descriptors & 39.3 $\pm$ 2.5 & 19.7 $\pm$ 0.8 \\
Combined classical & 44.8 $\pm$ 1.9 & 20.2 $\pm$ 0.8 \\
Small supervised CNN & \textbf{64.4} $\pm$ 6.5 & 19.0 $\pm$ 1.6 \\
\bottomrule
\end{tabular}
\end{table}

CROMA exceeds the CNN mean by 1.6 percentage points, with an inconclusive exploratory paired
comparison. Across twelve model comparisons against raw intensity statistics and the CNN, every
Holm-adjusted p-value remains above the 0.05 threshold. Establishing either equivalence or general
superiority would require stronger evidence.

\subsubsection{Classifiers trained on Earth}

\begin{figure*}[htbp]
\centering
\includegraphics[width=\textwidth]{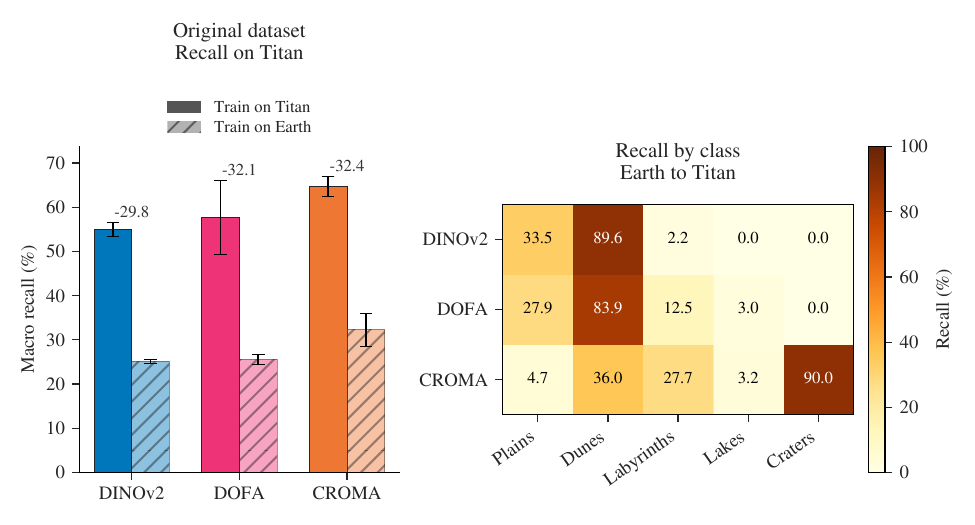}
\caption{Linear classifiers with encoder weights unchanged in the original dataset. Left: macro
recall after Titan or Earth training, evaluated on the five classes present in Earth. Transfer using
pretrained models stays above the 20\% five-class benchmark and is 29.8 to 32.4 percentage points
below Titan training. Right: transfer recall by class, excluding Hummocky because Earth supplies no
examples. Dunes has the highest mean transfer recall for DINOv2 and DOFA, at 89.6\% and 83.9\%.
CROMA's mean Craters recall is 90.0\%, with only 17 distinct test tiles across all folds. Error bars
show sample SD across five fold means after averaging classifier seeds. Identifying a physical
property that transfers between Earth and Titan would require evidence beyond recall.}
\label{fig:transfer}
\end{figure*}

Figure~\ref{fig:transfer} and Table~\ref{tab:cross_domain} compare classifiers trained on Earth
analog sites with classifiers trained on Titan, using the same five Titan test classes in both
cases. The six-class means in Table~\ref{tab:within_domain} answer a different question. Training
sets differ in size, planet, surface materials, sensor, frequency, polarization, incidence geometry,
native resolution, image product and curation. The transfer comparison therefore combines all of
these differences.
\begin{table}[t]
\centering
\caption{Classifiers trained on Earth and evaluated on the five Titan classes present in Earth
training. Values are mean $\pm$ sample standard deviation across five folds.
Linear-classifier values are first averaged over seeds in each fold; $k$-NN
gives one reproducible result per fold. $\Delta$ is the difference from the
corresponding classifier trained on Titan and evaluated on the same classes. The linear classifiers
trained on Earth and on Titan both retain six output logits. Untrained ViT remains near the
class-frequency benchmark despite its positive difference, providing no evidence of useful transfer.
CROMA has the highest transfer mean at 32.3\%, followed by DOFA at 25.5\% and DINOv2 at 25.1\%.
Comparisons require identical normalization, catalogs and image sampling. Bold indicates the highest
mean in each recall or kappa column for its classifier. Transfer differences remain in ordinary
type.}
\label{tab:cross_domain}
\begin{tabular}{lccccc}
\toprule
\textbf{Model} & \multicolumn{2}{c}{\textbf{Linear classifier}} & \textbf{$\Delta$ recall} & \multicolumn{2}{c}{\textbf{$k$-NN}} \\
\cmidrule(lr){2-3} \cmidrule(lr){5-6}
 & Macro recall & $\kappa$ & & Macro recall & $\kappa$ \\
\midrule
DINOv2 ViT-B/14       & 25.1 $\pm$ 0.4 & \textbf{0.135} $\pm$ 0.017 & $-$29.8 & 25.9 $\pm$ 1.0 & 0.146 $\pm$ 0.024 \\
DOFA ViT-B/16          & 25.5 $\pm$ 1.1 & 0.121 $\pm$ 0.014 & $-$32.1 & \textbf{26.1} $\pm$ 0.6 & \textbf{0.182} $\pm$ 0.018 \\
CROMA ViT-B/8          & \textbf{32.3} $\pm$ 3.7 & 0.034 $\pm$ 0.022 & $-$32.4 & 24.0 $\pm$ 2.5 & 0.094 $\pm$ 0.032 \\
Untrained ViT-B/14   & 20.0 $\pm$ 0.0 & 0.000 $\pm$ 0.000 & $+$3.6 & 20.2 $\pm$ 0.9 & 0.004 $\pm$ 0.020 \\
\bottomrule
\end{tabular}
\end{table}

Macro recall and $\kappa$ emphasize different aspects of the predictions. Among
the pretrained encoders, CROMA has the highest transfer macro recall, 32.3\%, and the lowest
$\kappa$, 0.034, compared with 0.135 for DINOv2 and 0.121 for DOFA. CROMA's mean
recall is 90.0\% for Craters, 36.0\% for Dunes, 4.7\% for Plains and 3.2\% for Lakes. Craters
contributes just 17 test tiles across all folds. Macro recall gives each class equal weight, whereas
$\kappa$ also depends on the frequencies of the labels and predictions.

\paragraph{Training-set size and image preparation:}
Table~\ref{tab:training_size_comparison} compares Earth training with Titan subsets of the same
total size. Complete Titan training folds contain 6{,}778 to 6{,}794 tiles; Earth supplies 352
training tiles and 46 validation tiles. The Titan classifiers still predict six classes, while
evaluation averages recall over the five classes present in Earth training. Stratified sampling
retains Titan class proportions subject to at least one tile per class. Uniform sampling imposes no
such class requirement. These choices equalize total training count while leaving class counts and
geographic diversity different from those of Earth.

\begin{table}[t]
\centering
\caption{Effect of training-set size on recall for the five Titan classes present in Earth. Every
training subset contains 352 tiles. Values are macro recall in percent, with sample standard
deviation across five fold means after averaging five seeds. The experiment equalizes total count
while allowing class composition and site diversity to differ. The values reproduce the workshop
training-size calculations. Bold indicates the highest mean recall across encoders for each training
setting.}
\label{tab:training_size_comparison}
\begin{tabular}{lccc}
\toprule
\textbf{Model} & \textbf{Earth} & \textbf{Titan stratified} & \textbf{Titan uniform} \\
\midrule
DINOv2 & 25.1 $\pm$ 0.4 & 43.9 $\pm$ 1.1 & 41.3 $\pm$ 2.1 \\
DOFA & 25.5 $\pm$ 1.1 & 45.9 $\pm$ 5.2 & 40.6 $\pm$ 3.8 \\
CROMA & \textbf{32.3} $\pm$ 3.7 & \textbf{51.1} $\pm$ 4.3 & \textbf{45.4} $\pm$ 4.4 \\
\bottomrule
\end{tabular}
\end{table}

With equal training counts, Titan's advantage is 13.1 to 20.4 percentage points. Reducing the Titan
sample also changes its class counts and geographic composition. The resulting decrease therefore
cannot be attributed solely to the number of labels or used to bound a planetary contribution. The
352-tile limit applies to classifier fitting. Image normalization for the saved Titan features still
uses the complete training fold and therefore uses more Titan training data. At the same number of
epochs, subsampling also changes class weights and the number of optimizer updates.

Three separate image transformations are evaluated: rendering Earth dB values as display DN,
quantizing normalized inputs, and smoothing the Earth output grid. Percentile rendering changes
DINOv2 and DOFA by less than one percentage point and reduces CROMA by 10.3 points. Quantization to
256 or 64 levels changes each model by at most 0.03 points. Gaussian smoothing with
$\sigma=0.65$ pixels changes recall by $-2.6$,
$-0.2$ and $-3.0$ points for DINOv2, DOFA and CROMA,
respectively. Combining smoothing and percentile rendering changes CROMA by
$+1.2$ points.
These transformations test sensitivity to the available image products. Smoothing an already
resampled grid cannot reconstruct native radar looks, and estimating display percentiles cannot
recover the USGS stretch. Every comparison leaves some differences in training count, class
composition, image products or site diversity. DOFA band identifiers in the independently generated
image-transformation arrays also require verification.
The smoothing scale and the Titan-range rendering variant reuse statistics from fold 1 training
images, including images tested in folds 3, 4 and 5. This use of test information limits those
experiments to exploratory interpretation. In addition, the standalone DN recall calculation and its
later MMD calculation estimate their display percentiles from different pixel samples, so they
represent different renderings. MMD also uses all Titan test classes and recalculates its bandwidth
for each condition and resample; recall excludes Hummocky. Population, rendering and kernel
differences must therefore be considered before relating changes in recall to changes in MMD.

\subsubsection{Earth and Titan feature distributions}
\label{sec:feature_distributions}

\begin{figure}[htbp]
\centering
\includegraphics[width=\columnwidth]{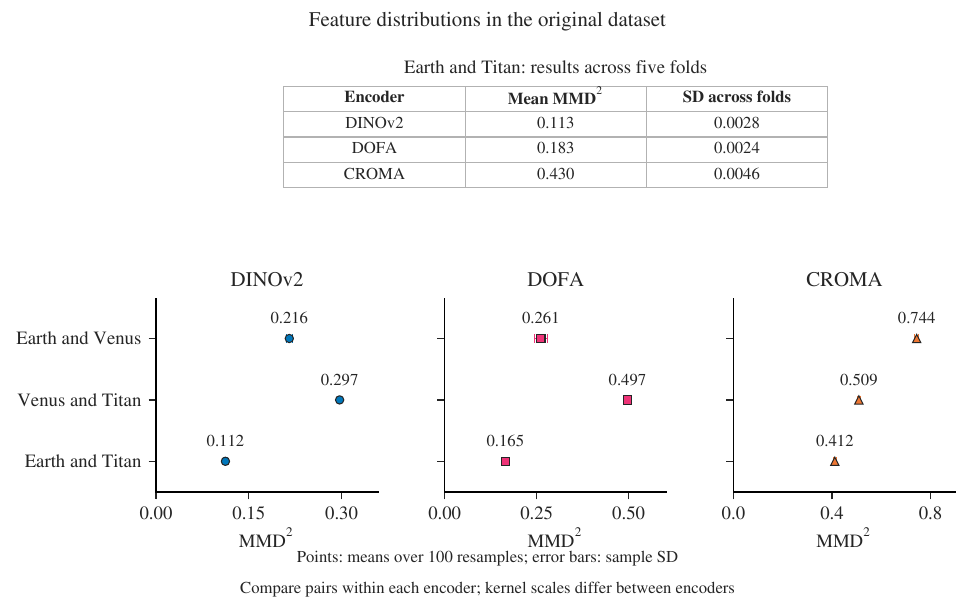}
\caption{Feature distributions in the original dataset. Top: Earth and Titan
MMD$^2$ from repeated equal-size samples, shown as mean and sample standard
deviation across five fold means for pretrained encoders. The random encoder is omitted from this
subplot. Stable arithmetic confirms its repeated estimate, but its original permutation statistic is
invalid. Bottom: MMD$^2$ for Earth, Venus and Titan, shown separately for
each encoder. Each encoder uses one bandwidth across domain pairs and equal sample counts in each
pair. Error bars give sample SD across 100 draws and describe subsampling variation. Earth and Titan
have the smallest estimates for the pretrained encoders under this estimator. Interpretation is
restricted to each model's feature space; physical causes and comparisons of magnitude between
models require further evidence.}
\label{fig:feature_distributions}
\end{figure}

Figure~\ref{fig:feature_distributions} and Table~\ref{tab:feature_distributions} compare Earth and
Titan features separately in each encoder's coordinate space.

The original Untrained ViT permutation calculation used float32 arithmetic and returned negative
observed statistics in four folds. Recalculation from the verified original arrays, using float64
direct squared distances, gives repeated biased MMD$^2$ of
$0.112 \pm 0.019$. These are the mean and sample standard deviation of five fold means;
each averages 100 resamples of 352 Earth and 352 Titan training tiles, with bandwidth estimated anew
for each resample. The repeated estimate is numerically valid, whereas the original permutation
statistic and the inference drawn from it are invalid.

The original constructor fallback and exact initial weights remain unverified, so the result applies
to those original arrays and cannot be assigned to the random encoder used with the enlarged
catalog. CETUS's reproduction documentation identifies the internal calculation and its access
limits. Weak classification is a separate result: low feature amplitude alone cannot explain RBF MMD
when bandwidth adapts to the features. Spatial exchangeability also remains an assumption after the
arithmetic is corrected.
Using equal domain counts, classifiers fitted to the same original arrays give proxy
$\mathcal{A}$-distances of $1.961 \pm 0.037$ for DINOv2,
$1.823 \pm 0.033$ for DOFA, $1.989 \pm 0.016$ for CROMA and
$1.516 \pm 0.114$ for the Untrained ViT. Values are means and sample standard deviations
over five folds. Every fit uses 352 tiles per domain, seed zero, a stratified half split,
standardization fitted on the classifier training subset, hinge-loss SGD and balanced test error.
Sampling, preprocessing and evaluation differ from the original calculation, and DOFA still receives
identifier 13.78 for both domains. These distances quantify domain discrimination under those
settings.

\begin{table}[t]
\centering
\caption{Earth and Titan feature comparisons for the original dataset, averaged over five geographic
folds. MMD$^2$ is the biased RBF-kernel V-statistic averaged over 100
equal-size resamples in each fold. The Untrained ViT estimate is recalculated from its original
arrays using stable arithmetic; its fold standard deviation is given in the text. The proxy
$\mathcal{A}$-distance column retains the original estimates with unequal domain
counts. The text gives the later estimates using equal domain counts under a different fitting
design. Centroid distances use the unnormalized coordinates of each model. These comparisons
characterize feature distributions; identifying physical causes of transfer error requires
additional evidence.}
\label{tab:feature_distributions}
\begin{tabular}{lccc}
\toprule
\textbf{Model} & \textbf{MMD$^2$} & \textbf{Original} $\boldsymbol{d_\mathcal{A}}$ & \textbf{Centroid Dist.} \\
\midrule
DINOv2 ViT-B/14       & 0.113 & 1.948 & 25.93 \\
DOFA ViT-B/16          & 0.183 & 1.918 & 2.32 \\
CROMA ViT-B/8          & 0.430 & 1.990 & 6.18 \\
Untrained ViT-B/14   & 0.112 & 1.817 & 0.002 \\
\bottomrule
\end{tabular}
\end{table}

\begin{figure*}[htbp]
\centering
\includegraphics[width=\textwidth]{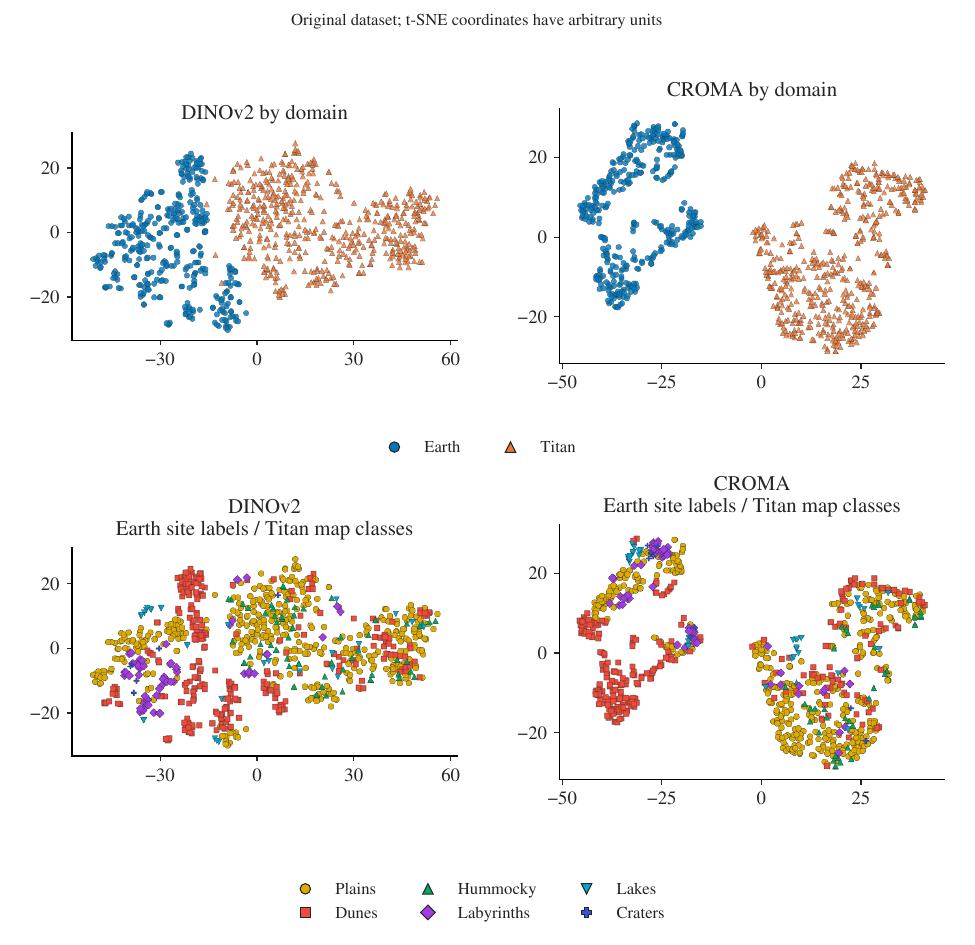}
\caption{t-distributed stochastic neighbor embedding (t-SNE) of original fold 1 features: DINOv2 at
left and CROMA at right. Each model uses all 352 Earth training tiles and 500 Titan test tiles. A
common random stream with seed 42 draws different Titan subsets for the models. The same coordinates
appear twice, colored by domain above and by Earth site label or Titan map class below. Both fits
use perplexity 30 and seed 42. The plotted coordinate systems are arbitrary. Separation or overlap
in the original feature spaces, and any relation to transfer performance, require measurements
beyond these projections.}
\label{fig:feature_embeddings}
\end{figure*}

\subsubsection{Sensitivity of MMD estimates}
\label{sec:mmd_sensitivity}

Table~\ref{tab:robustness} compares the Earth and Titan MMD estimates over five folds. For
pretrained encoders, the unbiased U-statistic, the mean after omitting Earth sites in turn, and the
mean from resampling Titan geographic regions differ from the primary biased estimate by at most
0.004. With 1{,}000 permutations, the test results round to $p=0.001$, near the
calculation's resolution limit. Their interpretation depends on the implemented exchangeability
assumptions; residual spatial dependence is unmeasured.
The Untrained ViT's repeated biased estimate agrees with the stable recalculation of its original
arrays. Its original permutation outputs are numerically invalid and cannot establish equality of
distributions. The numerical correction leaves the original catalog's geographic omission and
dependence between samples unchanged, so both continue to limit inference.

\begin{table}[htbp]
\centering
\caption{Sensitivity of Earth and Titan MMD$^2$ to estimation and resampling
choices. MMD$^2$ estimates are means across five Titan geographic folds. The
final column gives the largest permutation $p$-value. Each row uses that
model's feature space, which sets the scale and prevents direct comparisons of magnitude across
models.}
\label{tab:robustness}
\small
\begin{tabular}{lccccc}
\toprule
\textbf{Model} & \textbf{Biased} & \textbf{Unbiased} & \shortstack{\textbf{Omit each}\\\textbf{Earth site}} & \shortstack{\textbf{Resample Titan}\\\textbf{regions}} & \textbf{Max perm.\ $p$} \\
\midrule
DINOv2 & 0.113 & 0.110 & 0.115 & 0.113 & 0.001 \\
DOFA   & 0.183 & 0.180 & 0.182 & 0.183 & 0.001 \\
CROMA  & 0.430 & 0.427 & 0.430 & 0.431 & 0.001 \\
\bottomrule
\end{tabular}
\end{table}

\subsubsection{Recall by map class}

Figure~\ref{fig:feature_embeddings} projects the features into two dimensions using t-SNE.
Table~\ref{tab:per_class} reports recall for each class from linear classifiers trained on Titan or
Earth. Recall varies strongly by class. Separating the effects of morphology, material properties
and other differences between the image domains requires additional tests.

\begin{table}[t]
\centering
\caption{Recall by class for linear classifiers, in \%, averaged over seeds in each fold and then
over five geographic folds. Parentheses give the total number of distinct test tiles outside Selk;
each tile is tested once. N/A marks Hummocky because Earth supplies no training examples. Small
class counts make estimates unstable, especially for Craters, with $n=17$. Bold
indicates the highest mean recall among all four encoders for each class and training domain.}
\label{tab:per_class}
\resizebox{\textwidth}{!}{%
\begin{tabular}{lcccccccccccc}
\toprule
 & \multicolumn{2}{c}{\textbf{Plains} (7441)} & \multicolumn{2}{c}{\textbf{Dunes} (2215)} & \multicolumn{2}{c}{\textbf{Hummocky} (1014)} & \multicolumn{2}{c}{\textbf{Labyrinths} (257)} & \multicolumn{2}{c}{\textbf{Lakes} (362)} & \multicolumn{2}{c}{\textbf{Craters} (17)} \\
\cmidrule(lr){2-3} \cmidrule(lr){4-5} \cmidrule(lr){6-7} \cmidrule(lr){8-9} \cmidrule(lr){10-11} \cmidrule(lr){12-13}
\textbf{Model} & T$\to$T & E$\to$T & T$\to$T & E$\to$T & T$\to$T & E$\to$T & T$\to$T & E$\to$T & T$\to$T & E$\to$T & T$\to$T & E$\to$T \\
\midrule
DINOv2  & \textbf{71.2} & \textbf{33.5} & \textbf{76.0} & \textbf{89.6} & 61.8 & N/A & 50.1 & 2.2 & 75.3 & 0.0 & 1.6 & 0.0 \\
DOFA    & 39.7 & 27.9 & 63.5 & 83.9 & 60.6 & N/A & \textbf{67.1} & 12.5 & 81.7 & 3.0 & 36.0 & 0.0 \\
CROMA   & 57.6 & 4.7 & 70.8 & 36.0 & \textbf{72.8} & N/A & 67.0 & 27.7 & \textbf{83.4} & 3.2 & \textbf{44.7} & \textbf{90.0} \\
Untrained ViT  & 44.0 & 0.0 & 36.0 & 20.0 & 20.0 & N/A & 0.0 & \textbf{40.0} & 1.9 & \textbf{20.0} & 0.0 & 20.0 \\
\bottomrule
\end{tabular}}
\end{table}

For training and testing on Titan, linear classifiers give higher macro recall than cosine
$k$-NN for all three pretrained encoders under these settings. This
classifier comparison is insufficient to establish linear separability or the local structure of the
classes. After Earth training, CROMA gives the highest linear-classifier mean but the lowest
pretrained $k$-NN mean, 24.0\%, showing that the encoder ranking depends on
the classifier.

With Titan training, DINOv2 has the highest recall for Plains and Dunes, CROMA for Hummocky and
Lakes, and DOFA for Labyrinths. These differences describe agreement with mapped terrain classes.
Testing which physical image properties drive them would require a separate analysis.

After Earth training, Dunes gives the highest recall for DINOv2 at 89.6\% and DOFA at 83.9\%. Among
the pretrained encoders, CROMA gives the highest recall for Labyrinths and Craters, 27.7\% and
90.0\%. The Craters value depends on only 17 test tiles across all folds and is especially unstable.
Hummocky is omitted because Earth training contains none. Table~\ref{tab:per_class} presents recall
because its class average defines the original macro metric. Table~\ref{tab:transfer_metrics},
together with the Craters precision and F1 values, describes the false positives that recall alone
leaves out.

\subsubsection{Selk uncertainty and disagreement}

The Selk tiles are examined only after model development. They are excluded from every training,
validation and test fold. For each tile, linear classifiers trained on Titan produce class
probabilities and predictions from encoders whose weights remain unchanged. The ensemble comprises
$N=15$ combinations of pretrained model and geographic fold: three models
across five folds. Entropy is calculated from the mean probability vector,
$H(\bar{\mathbf{p}})=-\sum_{c=1}^{6}\bar{p}_c\log\bar{p}_c$,
and disagreement as $D=1-n_{\max}/N$, where $n_{\max}$ is the number of
members giving the most frequent class prediction. The Untrained ViT and the VIMS display values are
excluded.

Entropy and disagreement are each scaled to $[0,1]$ using their own minimum and
maximum among the 65 Selk tiles. The arithmetic mean of the two scaled quantities ranks the tiles
within this set of models and images. A high score may arise from ambiguous map classes, transitions
between units, image defects, model limitations or other unresolved causes. The ranking describes
variation in classifier uncertainty and disagreement. Using it for anomaly detection, calibrated
confidence or risk, geological prioritization or mission operations would require independent
validation.

\begin{figure}[htbp]
\centering
\includegraphics[width=\columnwidth]{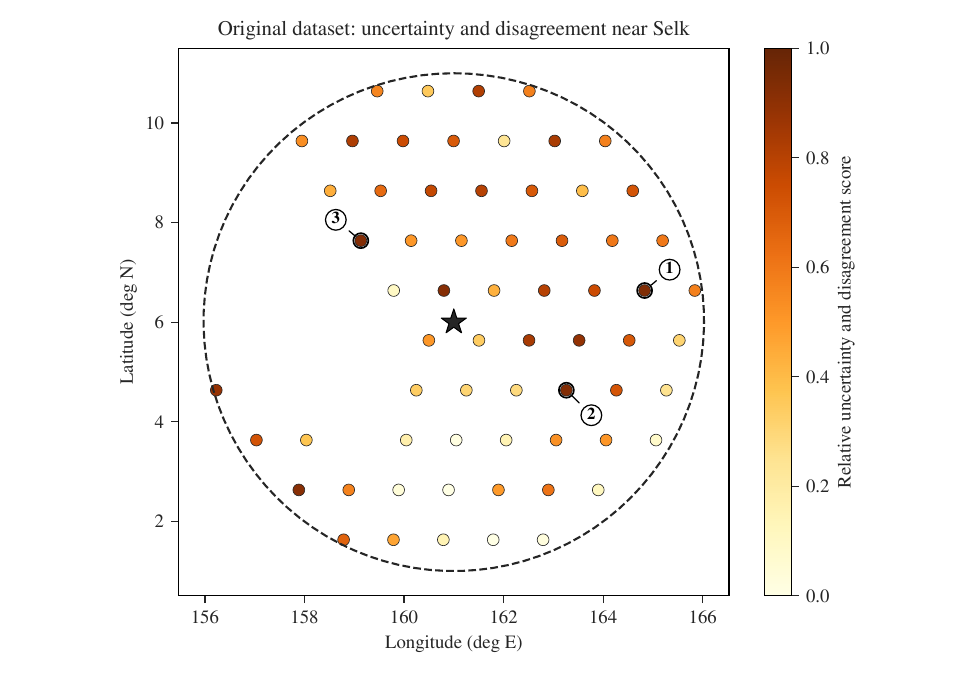}
\caption{Model uncertainty and disagreement for the original 65 Selk tiles. The star is the center
at 6\textdegree{}N, 161\textdegree{}E; the dashed curve lies 5\textdegree{} from it on a sphere.
Color shows the mean of entropy and disagreement after each is scaled separately to
$[0,1]$ using its minimum and maximum over these 65 tiles. Numbered rings mark
the three tiles with the highest scores in rank order. The first is the tile at 6.64\textdegree{}N,
164.83\textdegree{}E. Scores rank only this set of images and models. Interpreting them as
calibrated probabilities of anomaly, novelty, scientific value, risk or mission priority would
require independent validation.}
\label{fig:selk_uncertainty}
\end{figure}

The 65 tiles reserved near Selk lie within 5\textdegree{} of 6\textdegree{}N, 161\textdegree{}E.
This area overlaps the Selk exploration region examined in Dragonfly landing-site
studies~\citep{lorenz2021landingsite,bonnefoy2022selk}. Figure~\ref{fig:selk_uncertainty} maps the
relative ranking obtained by combining entropy and model disagreement for expert inspection.

The map labels the 65 Selk tiles as 37 Dunes, 17 Plains, 8 Craters and 3 Hummocky. The 15-member
ensemble, comprising three models in five fold-dependent feature spaces, predicts 34 Dunes, 13
Plains, 17 Hummocky and 1 Crater. Mean entropy is 0.963 against a maximum of
$\ln 6 = 1.79$, mean disagreement is 0.250 and mean relative score is 0.525. These
statistics describe the ensemble's predictions. Interpreting them as geological transitions or
operational priorities requires separate evidence.

\subsubsection{Transforming features and training encoder layers}

Two ways of using Titan information are compared: CORAL uses unlabeled Titan features to transform
Earth features, while supervised training of the final transformer blocks uses Titan labels.

\begin{samepage}
\paragraph{CORAL feature transformation:}
CORAL~\citep{sun2016coral} aligns Earth feature statistics with those of Titan. The transformation
subtracts the Earth mean, transforms the covariances, then adds the Titan mean:
$F_{\text{aligned}} = (F - \mu_{\text{src}}) \cdot C_{\text{src}}^{-1/2} C_{\text{tgt}}^{1/2} + \mu_{\text{tgt}}$.
The transformation is applied to Earth features before fitting the linear classifier; encoder
weights remain unchanged. The sample covariance matrices $C_{\text{src}}$ and
$C_{\text{tgt}}$ each receive $10^{-5}I$. Eigenvalues are floored at
$10^{-5}$ before taking square roots and inverse square roots. Subtracting the
source mean is part of the original formulation; adding the target mean specifies the coordinates
used here. Table~\ref{tab:coral} compares classification before and after this transformation.

\end{samepage}

\begin{table}[t]
\centering
\caption{Earth-to-Titan classification before and after CORAL, measured by macro recall in percent
over five classes. Metrics average seeds in each fold and then the five folds. CORAL uses unlabeled
Titan training features and evaluates classification on separate test regions. Both
MMD$^2$ calculations use the fitting samples and estimate bandwidth
separately before and after the transformation. Their ratio describes those changing kernels on
training features; measuring a reduction for one kernel on independent data would require a separate
calculation. Bold indicates the higher mean recall before or after CORAL for each encoder. MMD
values remain in ordinary type.}
\label{tab:coral}
\begin{tabular}{lccc}
\toprule
\textbf{Model} & \textbf{Before CORAL} & \textbf{CORAL} & \textbf{MMD$^2$ before/after} \\
\midrule
DINOv2 & \textbf{25.1} & 23.1 & 0.113 / 0.009 \\
DOFA   & 25.5 & \textbf{26.4} & 0.183 / 0.006 \\
CROMA  & 32.3 & \textbf{35.3} & 0.430 / 0.009 \\
\bottomrule
\end{tabular}
\end{table}

CORAL changes macro recall by $-2.0$ ($p=0.017$),
$+1.0$ ($p=0.370$) and $+2.9$
($p=0.417$) percentage points for DINOv2, DOFA and CROMA, respectively. These
exploratory paired tests use overlapping folds without adjustment for multiple comparisons. The
small MMD$^2$ after transformation comes from the same features used to
estimate CORAL, with bandwidth estimated again. Measuring a reduction of discrepancy on independent
data would require test features and one kernel used for both measurements. The remaining
classification errors leave their class-dependent and physical causes undetermined. Aligning
physical radar intensity would require compatible calibrated products and accounting for acquisition
conditions.

\paragraph{Training the final two transformer blocks:}
The last two transformer blocks of each encoder are trained on Titan images.
Table~\ref{tab:finetuning} averages five training seeds in each geographic fold. For comparison,
linear classifiers are fitted again on features from encoders whose weights remain unchanged, using
the same folds and seeds, so their values differ slightly from Table~\ref{tab:within_domain}.
These experiments used the implementation before weighted gradient accumulation was corrected.
Table~\ref{tab:encoder_training_scores} reports results from the current implementation and enlarged
catalog. Because the dataset and training design also changed, comparing these scores cannot isolate
the effect of the correction.

\begin{table}[t]
\centering
\caption{Training the final two transformer blocks on the original Titan dataset for 50 epochs with
AdamW, cosine annealing and batch size 8. Values are mean $\pm$ sample
standard deviation across five geographic folds after averaging five training seeds per fold. These
runs used the implementation before weighted gradient accumulation was corrected.
Table~\ref{tab:encoder_training_scores} gives the current implementation on the enlarged catalog.
Bold indicates the higher mean for each encoder and metric between classifier training alone and
training the final two encoder blocks.}
\label{tab:finetuning}
\begin{tabular}{lcccc}
\toprule
 & \multicolumn{2}{c}{\textbf{Encoder weights unchanged}} & \multicolumn{2}{c}{\textbf{Train final 2 blocks}} \\
\cmidrule(lr){2-3} \cmidrule(lr){4-5}
\textbf{Model} & Recall & $\kappa$ & Recall & $\kappa$ \\
\midrule
DINOv2 & 55.9 $\pm$ 2.9 & 0.502 $\pm$ 0.038 & \textbf{60.5} $\pm$ 1.9 & \textbf{0.531} $\pm$ 0.041 \\
DOFA   & \textbf{57.8} $\pm$ 8.4 & \textbf{0.295} $\pm$ 0.031 & 30.6 $\pm$ 2.4 & 0.124 $\pm$ 0.012 \\
CROMA  & \textbf{66.1} $\pm$ 2.5 & 0.432 $\pm$ 0.038 & 62.4 $\pm$ 2.2 & \textbf{0.532} $\pm$ 0.036 \\
\bottomrule
\end{tabular}
\end{table}

\begin{figure*}[htbp]
\centering
\includegraphics[width=\textwidth]{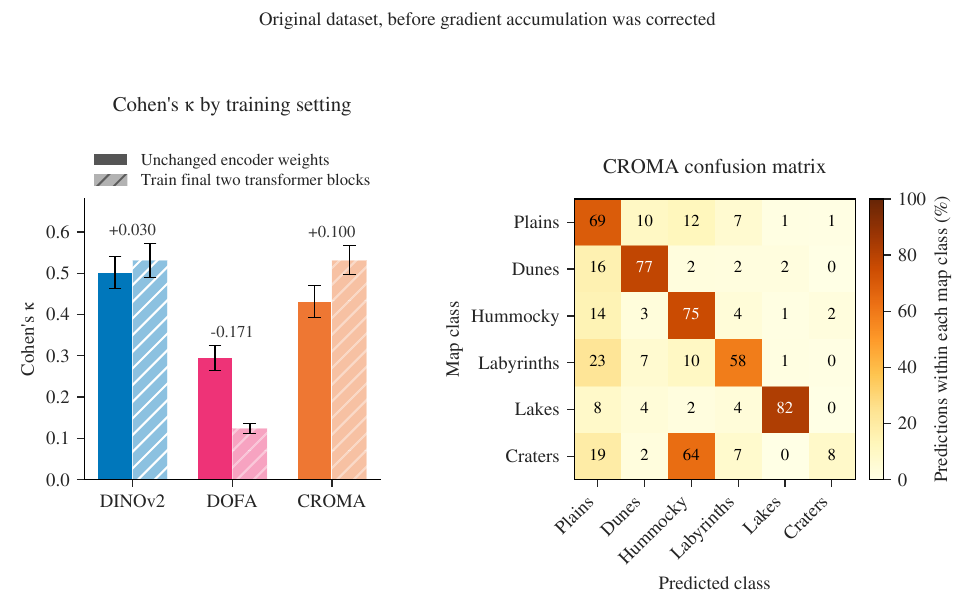}
\caption{Results from the original implementation, before weighted gradient accumulation was
corrected. Left: Cohen's $\kappa$ for linear classifiers using encoders whose
weights remain unchanged and for training the final two transformer blocks. Bars are means over five
geographic folds; error bars give the sample standard deviation of fold means. Right: CROMA
confusion matrix after training the two blocks, combining predictions over all folds and seeds
before normalizing each row. For CROMA, higher $\kappa$ together with lower macro
recall reflects a redistribution of errors. The combined counts describe the predictions, while
uncertainty calculations use fold means.}
\label{fig:finetuning}
\end{figure*}

Figure~\ref{fig:finetuning} shows the changes under the original implementation. DINOv2 gains 4.6
percentage points in macro recall and 0.030 in $\kappa$. CROMA loses 3.7
percentage points of macro recall while gaining 0.100 in $\kappa$, indicating a
redistribution of errors among classes. DOFA declines by 27.2 percentage points and 0.171 in
$\kappa$. These values apply to the original dataset and training code. The
enlarged catalog and corrected implementation are evaluated separately. Attributing the older
changes to architecture, conditioning, optimization or class imbalance would require comparisons
that vary those factors independently.

\subsubsection{Recall and labeled training-set size}
\begin{figure}[htbp]
\centering
\includegraphics[width=\columnwidth]{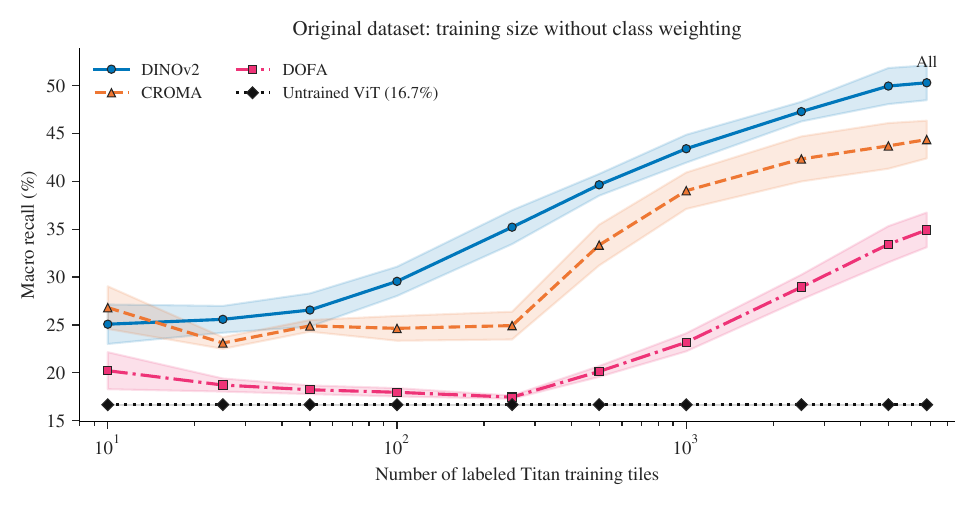}
\caption{Sensitivity to the number of labeled Titan training images in the original dataset, using a
loss without class weights. Lines give six-class macro recall averaged over five seeds for sampling
and training in each fold, then over five geographic folds. Bands show the sample standard deviation
across fold means. At $n=10$, CROMA has the highest mean, 26.8\% versus 25.1\%
for DINOv2; this difference is small relative to the fold spread. DINOv2 gives the highest result
using all training tiles, 50.3\%. All denotes each fold's complete training set, plotted at the mean
training size for DINOv2.}
\label{fig:label_efficiency}
\end{figure}
Figure~\ref{fig:label_efficiency} shows how macro recall changes as the number of labeled Titan
training tiles increases, with encoder weights unchanged and a linear classifier fitted to their
features. Stratified sampling guarantees at least one tile from each represented class. Small
subsets therefore contain disproportionately many rare-class examples and require labels before
images can be sampled. The curves apply to this sampling rule. Using them to forecast an annotation
campaign would require a design that accounts for how labels are acquired.
At 10 labels, CROMA reaches 26.8\%, DINOv2 25.1\% and DOFA 20.2\%, compared with the 16.7\%
six-class benchmark. Using every training tile in each fold, DINOv2 reaches 50.3\%, CROMA 44.4\% and
DOFA 34.9\%. DINOv2 improves most consistently as labels are added, while CROMA is nonmonotonic at
low counts. The loss here gives examples equal weight, whereas the experiment in
Table~\ref{tab:within_domain} uses class weights. These different losses can yield different scores
even when both experiments use every training tile.

\section{Image sources and Earth sites}
\label{app:dataset}

\subsection{Earth site details}

\begin{table}[H]
\centering
\caption{Earth site labels and intended morphological analogies. Assignments still need verification
for individual image areas, especially offshore Arctic and Icelandic areas and the site incorrectly
named Wadi Rum. Sentinel-1 IW VV products are processed with ESA SNAP and sampled at
\SI{351.5625}{\meter}/pixel.}
\small
\begin{tabular}{llll}
\toprule
\textbf{Terrain Class} & \textbf{Site Name} & \textbf{Location} & \textbf{Analog Rationale} \\
\midrule
\multirow{3}{*}{Dunes} & Namib Sand Sea & Namibia & Linear longitudinal dunes \\
 & Grand Erg Oriental & Algeria & Large-scale erg morphology \\
 & Rub' al Khali & Saudi Arabia & Massive sand sea \\
\midrule
\multirow{2}{*}{Craters} & Manicouagan & Canada & Large impact structure \\
 & Mistastin & Canada & Complex crater with melt sheet \\
\midrule
\multirow{2}{*}{Lakes} & Arctic thermokarst & Siberia & Shallow depressions \\
 & Salar de Uyuni & Bolivia & Low-relief salt flat \\
\midrule
\multirow{3}{*}{Labyrinths} & Atacama channels & Chile & Dissected plateau terrain \\
 & Wadi Rum label & Egypt, verified source & Labyrinths assignment needs review \\
 & Madagascar tsingy & Madagascar & Karst labyrinth terrain \\
\midrule
\multirow{3}{*}{Plains} & Arctic tundra & Canada & Low-relief terrain \\
 & Icelandic sandur & Iceland & Outwash plains \\
 & Desert pavement & California/Nevada, USA & Flat arid surface \\
\bottomrule
\end{tabular}
\end{table}

\subsection{Radar frequencies and image geometry}

\begin{table}[H]
\centering
\caption{Radar frequencies and benchmark geometry. Frequencies identify the instruments; the
experiments vary several factors together and leave the individual frequency effects undetermined.}
\small
\begin{tabular}{lrl}
\toprule
\textbf{Parameter} & \textbf{Value} & \textbf{Source} \\
\midrule
Cassini frequency & \SI{13.78}{\giga\hertz} & \citet{elachi2004cassini} \\
Cassini wavelength & \SI{2.176}{\centi\meter} & $c/f$ \\
Sentinel-1 frequency & \SI{5.405}{\giga\hertz} & ESA \\
Sentinel-1 wavelength & \SI{5.547}{\centi\meter} & $c/f$ \\
Titan sphere radius & \SI{2574.7}{\kilo\meter} & Benchmark geometry \\
Image width & \SI{45}{\kilo\meter} & Benchmark geometry \\
Common output spacing & \SI{351.5625}{\meter}/pixel & Benchmark geometry \\
\bottomrule
\end{tabular}
\end{table}

\section{Input configurations and computing resources}
\label{app:implementation}

Table~\ref{tab:model_config} specifies the encoders and inputs used with the original dataset.
Section~\ref{sec:experiment_design} gives the settings for the current evaluation. Computing
resources are described below for the enlarged Titan experiments.

\subsection{Model weights and input handling}

\begin{table}[H]
\centering
\caption{Model configurations used with the original dataset. The architecture chosen by the
Untrained ViT fallback and its exact initial weights remain unverified; see
Section~\ref{sec:model_descriptions}.}
\label{tab:model_config}
\small
\begin{tabular}{@{}p{0.70in}p{0.40in}p{0.88in}p{2.35in}p{1.10in}@{}}
\toprule
\raggedright Model & \raggedright Patch size & \raggedright Encoder channels & \raggedright Supplied SAR input & \raggedright Weights source \tabularnewline
\midrule
\raggedright DINOv2 & \raggedright 14 & \raggedright 3, RGB & \raggedright One SAR channel repeated three times & \raggedright Meta AI \tabularnewline
\raggedright DOFA & \raggedright 16 & \raggedright Variable & \raggedright One SAR channel; original extraction used 13.78 for Earth and Titan & \raggedright earthflow/DOFA \tabularnewline
\raggedright CROMA & \raggedright 8 & \raggedright 2, VV and VH & \raggedright One SAR channel repeated twice; $120\times120$ pixels & \raggedright CROMA authors \tabularnewline
\raggedright Untrained ViT & \raggedright 14 & \raggedright 1 & \raggedright One SAR channel and a single-channel stem & \raggedright Seed 42 \tabularnewline
\bottomrule
\end{tabular}
\end{table}

\subsection{Computing resources}

Feature extraction and encoder training for the enlarged Titan catalog ran on three NVIDIA RTX 3080
Ti GPUs, each with 12\,GB of memory. Independent jobs ran on separate GPUs. Linear classifiers and
$k$-NN ran on CPU using saved features.

\end{document}